\documentclass[runningheads]{llncs}
\usepackage{makeidx}
\usepackage{graphicx}
\usepackage{amsmath,amssymb} 
\usepackage{color}
\usepackage{comment}

\usepackage{color}
	
\usepackage[export]{adjustbox}
\usepackage{pifont}
\usepackage{algorithm}
\usepackage{algpseudocode}
\usepackage{xpatch}
\usepackage{multirow}
\algrenewcommand\alglinenumber[1]{{\sffamily\scriptsize#1}}

\makeatletter
\xpatchcmd{\algorithmic}{\itemsep\z@}{\itemsep=2ex plus2pt}{}{}

\makeatother

\newcommand{\etal}{{\it et al}.~}
\def\etc{{\it etc.}}
\def\eq{{\it eq.}}
\newcommand{\cmark}{\ding{51}}
\newcommand{\xmark}{\ding{55}}
\begin{document}

\pagestyle{headings}
\mainmatter
\def\ECCVSubNumber{59}

\title{Towards Bounding-Box Free\\Panoptic Segmentation}

\author{Ujwal Bonde$^{1}$ \quad
        Pablo F. Alcantarilla$^{1}$ \quad
        Stefan Leutenegger$^{1,2}$\quad
{\tt\small {firstname}@slamcore.com}\quad
}
\institute{$^1$ SLAMcore Ltd.\quad
$^2$ Imperial College London\\}
\maketitle

\begin{abstract} 
In this work we introduce a new Bounding-Box Free Network (BBFNet) for panoptic segmentation. 
Panoptic segmentation is an ideal problem for proposal-free methods as it already requires per-pixel 
semantic class labels. We use this observation to exploit class boundaries from off-the-shelf 
semantic segmentation networks and refine them to predict instance labels. Towards this goal BBFNet 
predicts coarse watershed levels and uses them to detect large instance candidates where boundaries 
are well defined. For smaller instances, whose boundaries are less reliable, BBFNet also predicts 
instance centers by means of Hough voting followed by mean-shift to reliably detect small objects. A 
novel triplet loss network helps merging fragmented instances while refining boundary pixels. Our 
approach is distinct from previous works in panoptic segmentation that rely on a combination of a 
semantic segmentation network with a computationally costly instance segmentation network based on 
bounding box proposals, such as Mask R-CNN, to guide the prediction of instance labels using a 
Mixture-of-Expert (MoE) approach. We benchmark our proposal-free method on Cityscapes and Microsoft 
COCO datasets and show competitive performance with other MoE based approaches while outperforming 
existing non-proposal based methods on the COCO dataset. We show the flexibility of our method using 
different semantic segmentation backbones.
\end{abstract}

\section{Introduction}\label{sec:intro}
Panoptic segmentation is the joint task of predicting semantic scene segmentation together with 
individual instances of objects present in the scene. Historically this has been explored under 
different umbrella terms of scene understanding~\cite{Yao12cvpr} and scene 
parsing~\cite{Tighe14cvpr}. In~\cite{Kirillov19cvprb}, Kirillov~\etal coined the term and gave a 
more concrete definition by including the suggestion from Forsyth~\etal~\cite{Forsyth96IWORCV} of 
splitting the objects categories into \textit{things} (countable 
objects like persons, cars, \etc) and \textit{stuff} (uncountable like sky, road, \etc) classes. 
While \textit{stuff} classes require only semantic label prediction, \textit{things} need both the 
semantic and instance labels. Along with this definition, \textit{Panoptic Quality} (PQ) 
measure was proposed to benchmark different methods. Since then, there has been a more focused 
effort towards panoptic segmentation with multiple 
datasets~\cite{Cordts16cvpr,Lin14eccv,Neuhold17iccv} supporting it.

\begin{figure}[t!]
\centering
  \setlength{\tabcolsep}{0.5pt}  
  \tiny{
  \begin{tabular}{cccc}
  \includegraphics[width=0.24\columnwidth, cfbox=blue 1pt 0pt]{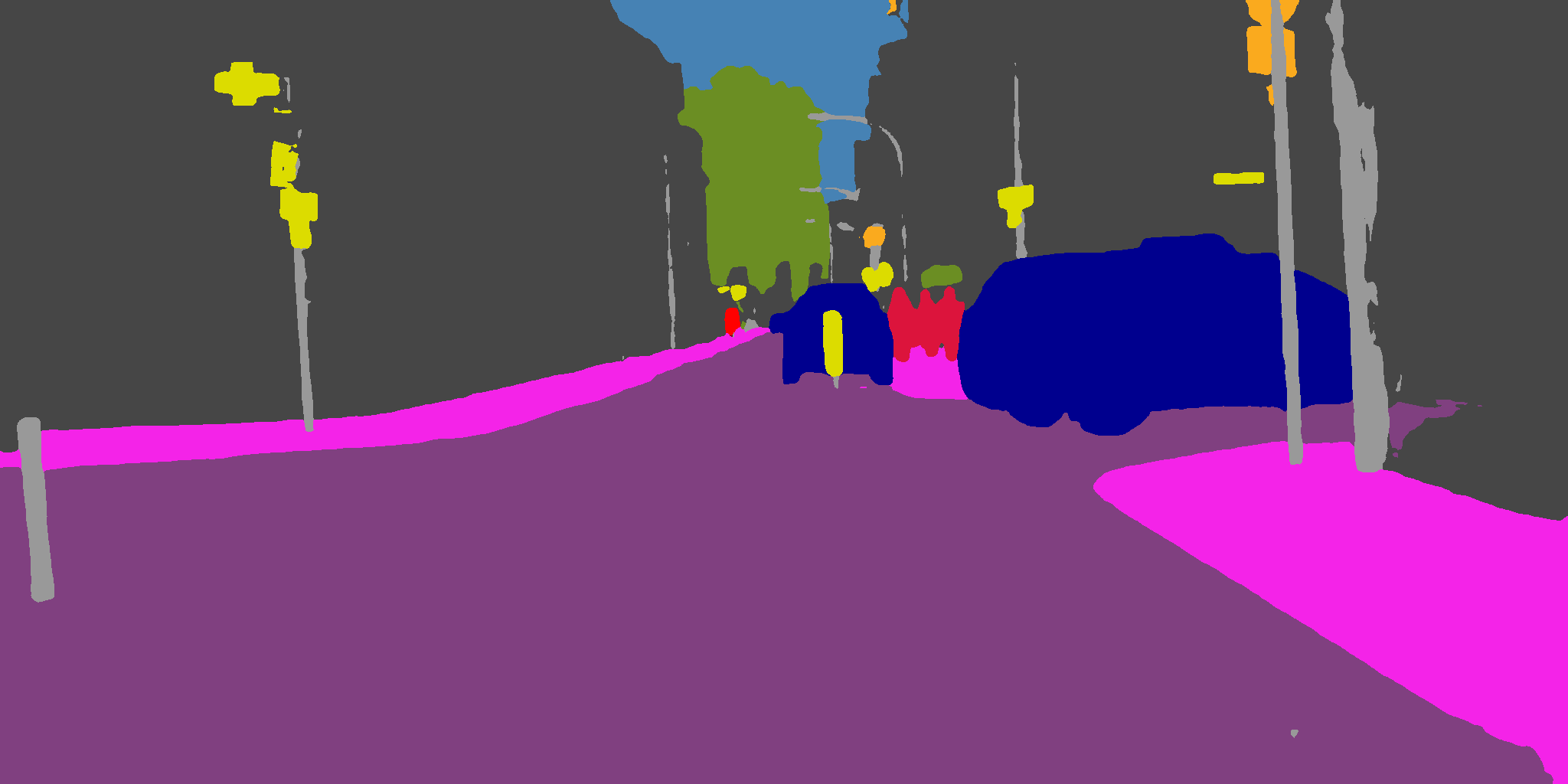}&
  \includegraphics[width=0.24\columnwidth, cfbox=blue 1pt 
0pt]{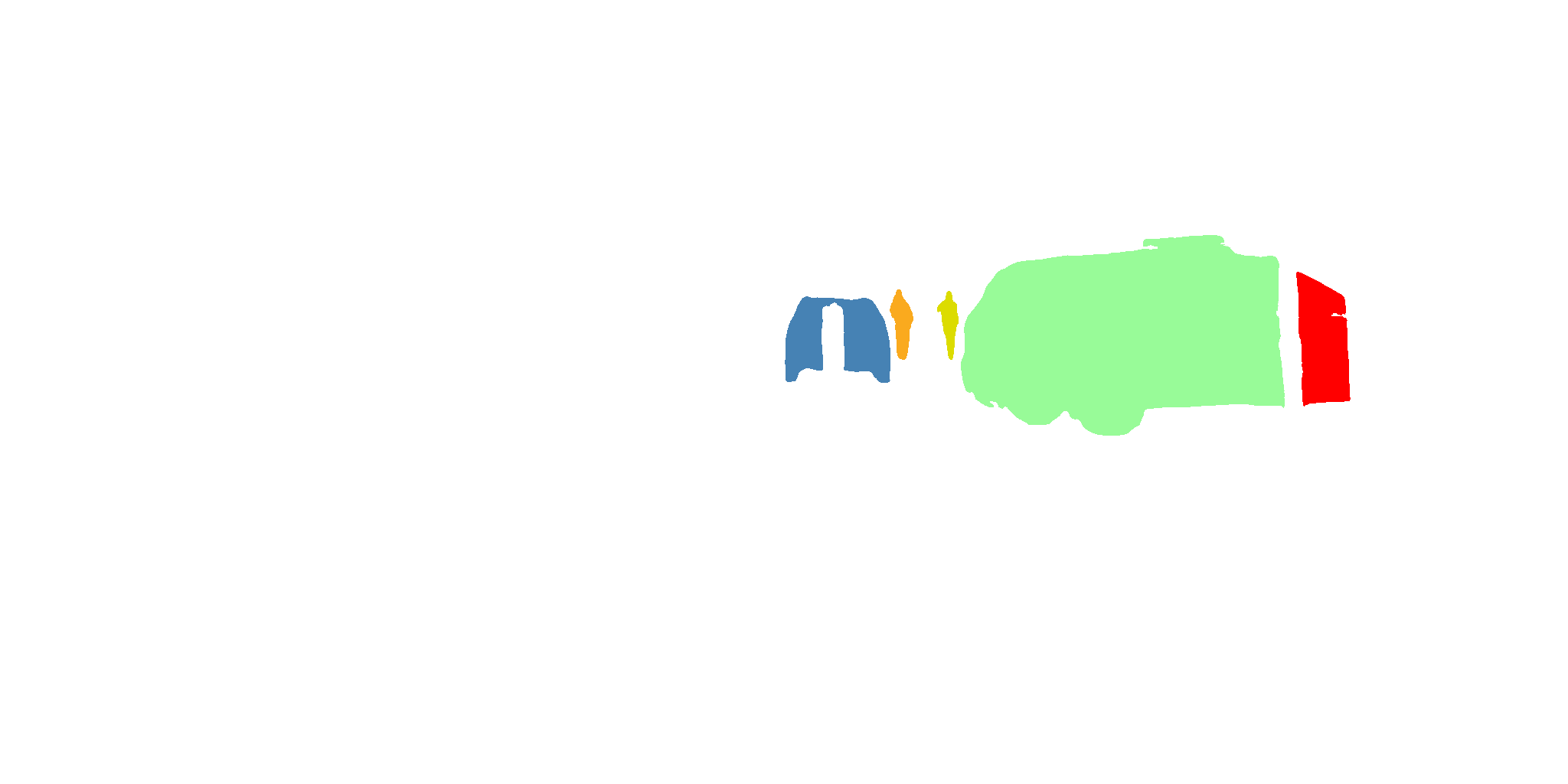} & 
\includegraphics[width=0.24\columnwidth, cfbox=blue 1pt 0pt]{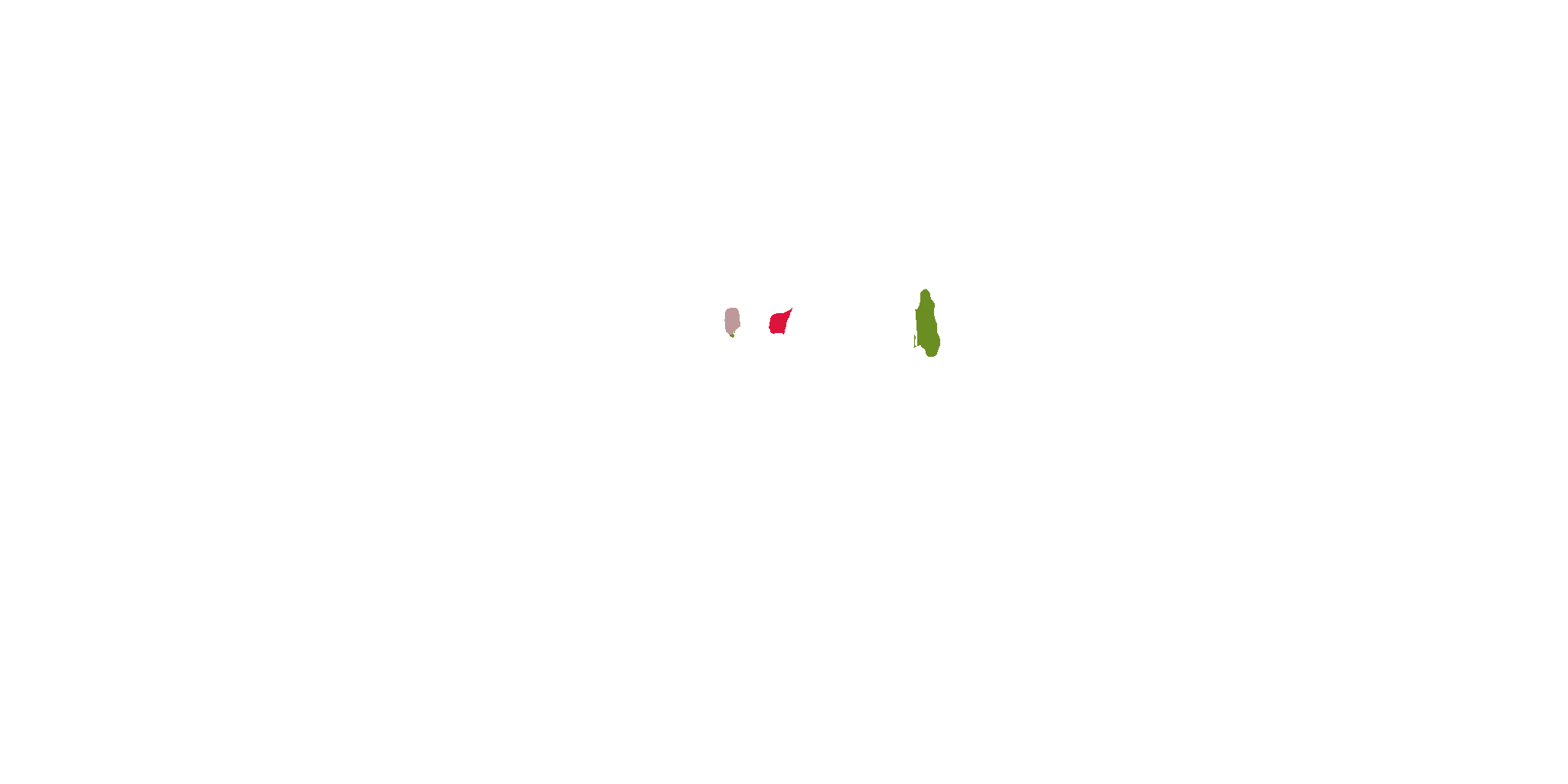}&
  \includegraphics[width=0.24\columnwidth, cfbox=blue 1pt 0pt]{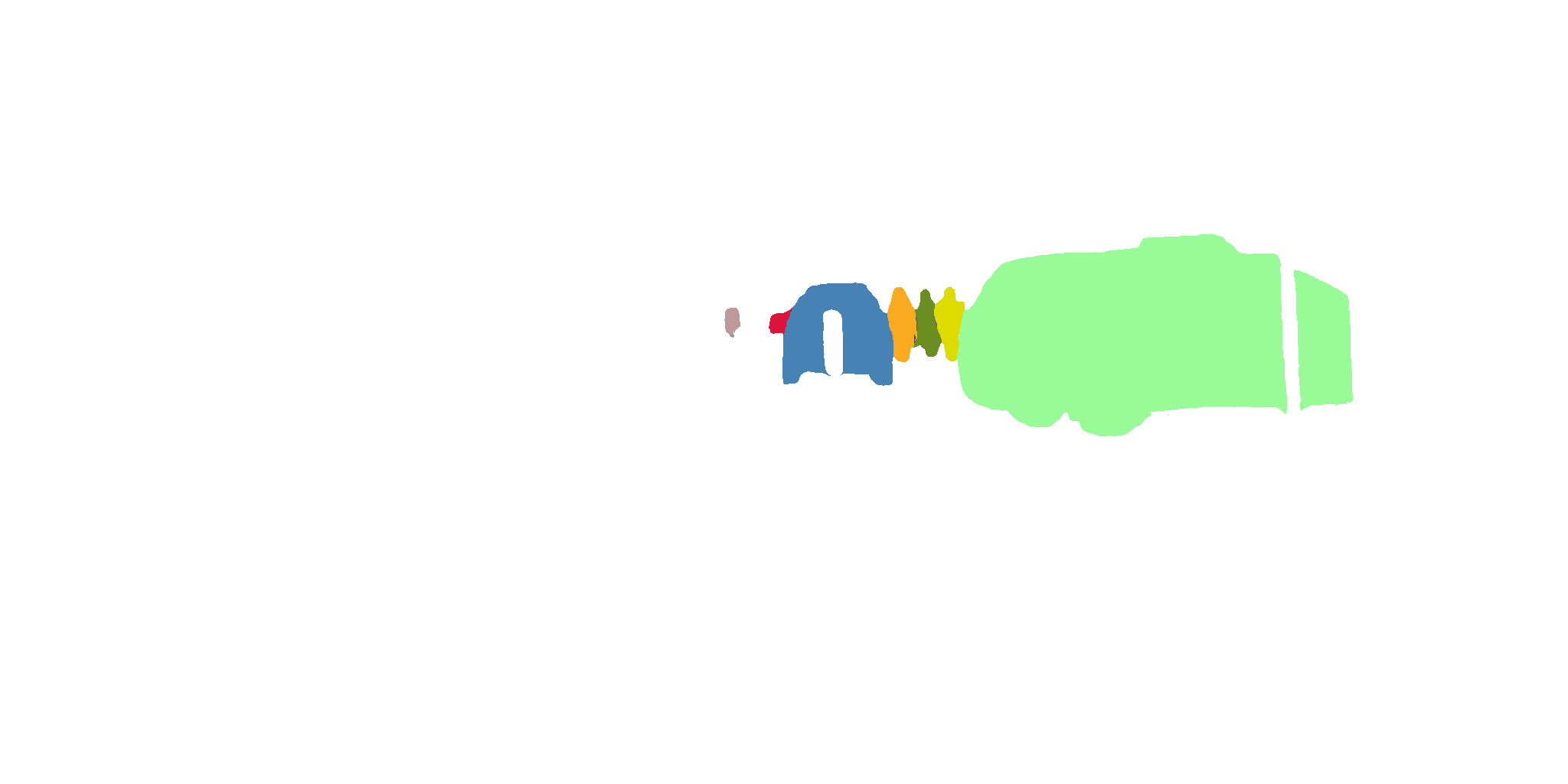}\\
  (a) S. Segmentation Head & (b) Watershed Head & (c) Hough Voting Head & (d) Triplet Loss 
Head\\
\end{tabular}}
\caption{BBFNet gradually refines the class boundaries of the semantic segmentation network to 
predict panoptic segmentation. The Watershed head detects candidates for large instances whereas 
the Hough voting head detects small object instances. The Triplet Loss network refines and merges 
the detection to obtain the final instance labels.}\label{fig:teaser}
\noindent\hrulefill 
\vspace{-0.5cm}
\end{figure}

Existing methods for panoptic segmentation can be broadly classified into two groups. The first 
group uses a proposal based approach for predicting \textit{things}. Traditionally these methods use 
completely separate instance and scene segmentation networks. Using a MoE approach, the outputs are 
combined either heuristically or through another sub-network. Although, more recent works propose 
sharing a common feature backbone for both networks~\cite{Kirillov19cvpra,Porzi19cvpr}, this split 
of tasks restricts the backbone network to the most complex branch. Usually this restriction is 
imposed by the instance segmentation branch, performed by Mask-RCNN~\cite{He17iccv}.

The second group of work uses a proposal free approach for instance segmentation allowing for a more 
efficient design. An additional benefit of these methods is that they do not need 
bounding-box predictions. While bounding-box detection based approaches have been popular and 
successful, they require predicting auxiliary quantities like scale, width and height which do not 
directly contribute to instance segmentation. Furthermore, the choice of bounding-boxes for 
object-detection had been questioned in the past~\cite{Redmon18arxiv}. We believe panoptic 
segmentation to be an ideal problem for a bounding-box free approach since it already contains 
structured information from semantic segmentation.

In this work, we exploit this using a flexible panoptic segmentation head that can be added to any 
off-the-shelf semantic segmentation network. We coin this as \textit{Bounding-Box Free Network} 
(BBFNet) which is a proposal free network and predicts \textit{things} by gradually refining the 
class boundaries predicted by the base network. To achieve this we exploit previous works in 
non-proposal based methods for instance 
segmentation~\cite{Bai17cvpr,Brabandere17arxiv,Neven17arxiv}. Based on the output of a semantic 
segmentation network, BBFNet first detects noisy and fragmented large instance candidates using a 
watershed-level prediction head (see Fig.~\ref{fig:teaser}). These candidate regions are clustered 
and their boundaries improved with a triplet loss based head. The remaining smaller instances, with 
unreliable boundaries, are detected using a Hough voting head that predicts the offsets to the 
center of the instance. Mean-shift clustering followed by 	vote back-tracing is used to reliably 
detect the smaller instances. Without using MoE our method produces comparable results to proposal 
based approaches while outperforming proposal-free methods on the COCO dataset. 

To summarise, we present BBFNet for panoptic segmentation which is a flexible, bounding-box free 
non-MoE approach that does not use the output of any instance segmentation or detection network 
while outperforming existing non-proposal based methods.

\section{Related Work}\label{sec:literature}
Despite the recent introduction of panoptic segmentation, there have already been multiple works 
attempting to address this~\cite{Geus19arxiv,Li19arxiv,Li19cvpr,Xiong19cvpr}. This is in part due 
to its importance to the wider community, success in individual subtasks of instance and semantic 
segmentation and publicly available datasets to benchmark different methods.

\noindent \textbf{Panoptic Segmentation:}
Most current works in panoptic segmentation fall under the proposal based approach for detecting 
\textit{things}. In~\cite{Kirillov19cvprb}, Kirillov~\etal use separate networks for semantic 
segmentation (\textit{stuff}) and instance segmentation (\textit{things}) with a heuristic MoE 
fusion of the two results for the final prediction. Realising the duplication of feature extractors 
in the two related tasks,~\cite{Kirillov19cvpra,Li19arxiv,Li19cvpr,Porzi19cvpr,Xiong19cvpr} propose 
using a single backbone feature extractor network. This is followed by separate branches for the two 
sub-tasks with a heuristic or learnable MoE head to combine the results. While panoptic Feature 
Pyramid Networks (FPN)~\cite{Kirillov19cvpra} uses Mask R-CNN~\cite{He17iccv} for the 
\textit{things} classes and fills in the \textit{stuff} classes using a separate FPN branch, 
UPSNet~\cite{Xiong19cvpr} combines the resized logits of the two branches to predict the final 
output. In AUNet~\cite{Li19cvpr}, attention masks predicted from the Region Proposal Network (RPN) 
and the instance segmentation head help fusing the results of the two tasks. Instead of relying only 
on the instance segmentation branch, TASCNet~\cite{Li19arxiv} predicts a coherent mask for the 
\textit{things} and \textit{stuff} classes using both branches. This is later filled with the 	
respective outputs. All these methods rely on Mask R-CNN~\cite{He17iccv} for predicting 
\textit{things}. Mask R-CNN is a two-stage instance segmentation network which uses a RPN to predict 
initial candidates for instance. The proposed candidates are either discarded or refined and a 	
separate head produces segmentation for the remaining candidates. The two-stage serial approach 
makes Mask R-CNN accurate albeit computationally expensive and inflexible thus slowing progress 
towards real-time panoptic segmentation. 

In FPSNet~\cite{Geus19arxiv}, the authors replace Mask R-CNN with a computationally less expensive 
detection network and use its output as a soft attention mask to guide the prediction of 
\textit{things} classes. This trade off is at a cost of considerable reduction in accuracy while 
continuing to use a computationally expensive backbone (ResNet50~\cite{He15cvpr}). 
In~\cite{Li20cvpr} the authors make up for the reduced accuracy by using an affinity network but 
this is at the cost of computational complexity. Both these methods still use bounding-boxes for 
predicting \textit{things}. In~\cite{Sofiiuk19iccv}, the detection network is replaced with an 
object proposal network which predicts instance candidates. In contrast, we propose a flexible 
panoptic segmentation head that relies only on a semantic segmentation network which, when replaced 
with faster networks~\cite{Romera18its,Sandler18cvpr} allows for a more efficient solution.

A parallel direction gaining increased popularity is the use of proposal-free approach for 
predicting \textit{things}. In~\cite{Uhrig16gcpr}, the authors predict the direction to 
the center and replace bounding box detection with template matching using these predicted 
directions as a feature. Instead of template matching,~\cite{Arnab17cvpr,Li18eccv} use a dynamically 
initiated conditional random field graph from the output of an object detector to segment 
instances. In the more recent work of Gao~\etal\cite{Gao19iccv}, cascaded graph partitioning is 
performed on the predictions of a semantic segmentation network and an affinity pyramid computed 
within a fixed window for each pixel. Cheng~\etal\cite{Cheng20cvpr} simplify this process by 
adopting a parallelizable grouping algorithm for \textit{thing} pixel to predict instance 
segmentation. In comparison, our flexible panoptic segmentation head predicts \textit{things} by 
refining the segmentation boundaries obtained from any backbone semantic segmentation network. 
Furthermore, our post-processing steps are computationally more efficient compared to other 
proposal-free approaches while outperforming them on multiple datasets.

\noindent \textbf{Instance segmentation:}
Traditionally predicting instance segmentation masks relied on obtaining rough boundaries followed 
by refining them~\cite{Ladicky09cvpr,Shotton09pami}. With the success of deep neural networks in 
predicting object proposals~\cite{Redmon16cvpr,Ren15nips}, and the advantages of an end-to-end 
learning method, proposal based approaches have become more popular. Recent works 
have suggested alternatives to predicting proposals in an end-to-end trainable network. As 
these are most relevant to our work, we only review these below. 

In~\cite{Bai17cvpr}, the authors propose predicting quantised watershed 
energies~\cite{Vincent91pami} using a Deep Watershed Transform network (DWT). Connected-components 
on the second-lowest watershed energy level are used to predict the instance segments. While this 
does well on large instances it suffers on small and thin instances. Moreover, fragmented regions 
of occluded objects end up being detected as different instances. In 
comparison,~\cite{Brabandere17arxiv} embed the image into a transformed feature space where pixels 
of the same instance cluster together and pixels of different instances are pushed apart. While 
this method is not affected by object fragmentation, poor clustering often leads to either 
clustering multiple objects as single instance (under-segmentation) or segmenting large objects 
into multiple instances (over-segmentation). In~\cite{Neven19cvpr}, the authors try to address this 
by using variable clustering bandwidths predicted by the network. In this work, we observe the 
complementary advantages of these methods and exploit it towards our goal 
of an accurate, bounding-box free panoptic segmentation.\\


\section{Panoptic Segmentation}\label{sec:BBFNet}

\begin{figure*}[th]
\centering
\includegraphics[width=\linewidth]{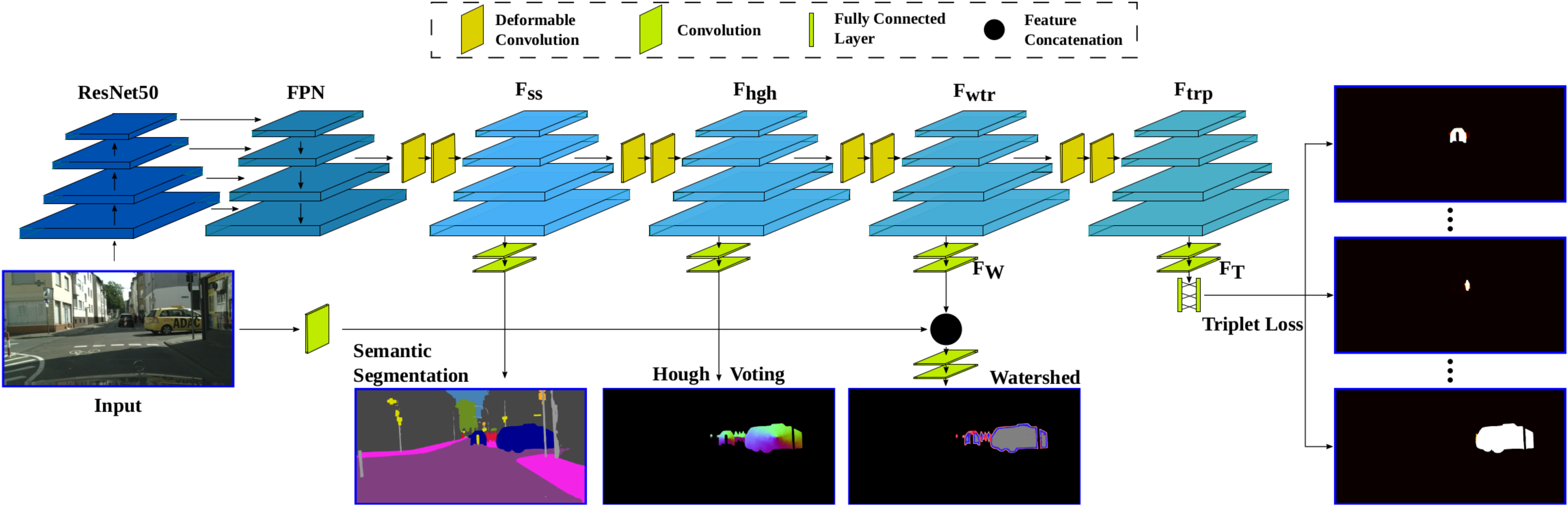}
\caption{BBFNet gradually refines the class boundaries of the backbone semantic segmentation 
network 
to predict panoptic segmentation. The watershed head predicts quantized watershed levels (shown in 
different colours) which is used to detect large instance candidates. For smaller instances we use 
Hough voting with fixed bandwidth. The output shows offsets ($X_{\text{off}}$, 
$Y_{\text{off}}$) colour-coded to represent the direction of the predicted vector. Triplet head 
refines and merges the detection to obtain the final instance labels. We show the class probability 
(colour-map \textit{hot}) for different instances with their center pixels used as $f_a$. 
Table~\ref{tab:BBFNet} lists the components of individual heads while $\S\ref{sec:BBFNet}$ 
explains them in detail.}\label{fig:method}
\noindent\hrulefill
\vspace{-0.5cm}
\end{figure*}

In this section we introduce our non-bounding box approach to panoptic segmentation. 
Fig.~\ref{fig:method} shows the various blocks of our network and Table~\ref{tab:BBFNet} details 
the main components of BBFNet. The backbone semantic segmentation network consists of a 
ResNet50 followed by an FPN~\cite{Lin17cvpr}. In FPN, we only use the P2, P3, P4 and P5 feature 
maps which contain $256$ channels each and are $1/4$, $1/8$, $1/16$ and $1/32$ of the original 
scale respectively. Each feature map then passes through the same series of eight Deformable 
Convolutions(DC)~\cite{Dai17iccv}. Intermediate features after every couple of DC are used to 
predict semantic segmentation ($\S\ref{sec:semanticSeg}$), Hough votes 
($\S\ref{sec:hough}$), watershed energies ($\S\ref{sec:watershed}$) and features for the triplet 
loss~\cite{Weinberger09jmlr} network. We first explain each of these components and their 
corresponding training loss. In ($\S\ref{sec:training}$) we explain our training and inference 
steps. Through ablation studies we show the advantages of each block in ($\S\ref{sec:ablation}$). 

\subsection{Semantic Segmentation}\label{sec:semanticSeg}
 
The first head in BBFNet is used to predict semantic segmentation. This allows BBFNet to 
quickly predict \textit{things} ($C_{\text{\textit{things}}}$) and \textit{stuff} 
($C_{\text{\textit{stuff}}}$) labels while the 
remainder of BBFNet improves \textit{things} boundaries using semantic segmentation features 
$F_{seg}$. We use per-pixel cross-entropy loss to train this head given by:
\begin{equation}\label{eq:sseg}
 L_{ss} = \sum_{c\in \{C_{\text{\textit{stuff}}},C_{\text{\textit{thing}}\}}} 
y_c\;\log(p_c^{ss}),
\end{equation}

\noindent where $y_c$ and $p_c^{ss}$ are respectively the one-hot ground truth label and 
predicted softmax probability for class $c$.

\subsection{Hough Voting}\label{sec:hough}
The Hough voting head is similar to the semantic segmentation head and is used to refine
 $F_{ss}$ to give Hough features $F_{hgh}$. These are then used to predict 
offsets for the center of each \textit{things} pixel. 
We use a $tanh$ non-linearity to squash the predictions and obtain normalised offsets 
($\hat{X}_{\text{off}}$ and $\hat{Y}_{\text{off}}$). Along with the centers we also predict the 
uncertainty in 
the two directions ($\sigma_x$ and $\sigma_y$) making the number of predictions from the Hough 
voting head equal to $4\times C_{\text{\textit{things}}}$. The predicted center for each pixel 
($x$, $y$), is then given by:
\begin{equation}
\begin{split}
  \hat{X}_{\text{\text{center}}}(x,y) = \hat{x}+\hat{X}_{\text{off}}^{C(x,y)}(x,y),\\
  \hat{Y}_{\text{\text{center}}}(x,y) = \hat{y}+\hat{Y}_{\text{off}}^{C(x,y)}(x,y),
 \end{split}
\end{equation}

\noindent where $C$ is the predicted class and ($\hat{x}$, $\hat{y}$) are image normalised pixel 
location. 

Hough voting is inherently noisy~\cite{Ballard81pr} and requires clustering or mode seeking methods 
like mean-shift~\cite{Cheng95pami} to predict the final object centers. As instances could have 
different scales, tuning clustering hyper-parameters is difficult. For this reason we use 
Hough voting primarily to detect small objects and to filter predictions from other heads. We 
also observe that the dense loss from the Hough voting head helps convergence of deeper heads in 
our network.

The loss for this head is only for the \textit{thing} pixels and is given by:
\begin{equation}\label{eq:hcent} 
 L_{hgh} = w\Big(\frac{(X_{\text{off}} - \hat{X}_{\text{off}})^2}{\sigma_x} + 
\frac{(Y_{\text{off}} 
-\hat{Y}_{\text{off}})^2}{\sigma_y}\Big)-\frac{1}{2} \Big(\log(\sigma_x) + \log(\sigma_y)\Big),
\end{equation}

\noindent where $X_{\text{off}}$ and $Y_{\text{off}}$ are ground truth offsets and $w$ 
is the per pixel weight. To avoid bias towards large objects, we inversely weigh the instances 
based on the number of pixels. This allows it to accurately predict the centers for objects of all 
sizes. Note that we only predict the centers for the visible regions of an instance and do not 
consider its occluded regions.

\subsection{Watershed Energies}\label{sec:watershed}
Our watershed head is inspired from DWT~\cite{Bai17cvpr}. Similar to that work, we quantise the 
watershed levels into fixed number of bins ($K=4$). The lowest bin ($k=0$) corresponds to 
background and regions that are within 2 pixels inside the instance boundary. Similarly, $k=1$, 
$k=2$ are for regions that are within 5 and 15 pixels away from the instance boundary, 
respectively, while $k=3$ is for the remaining region inside the instance. 

In DWT, the bin corresponding to $k=1$ is used to detect large instance boundaries. While this does 
reasonably well for large objects, it fails for smaller objects producing erroneous boundaries. 
Furthermore, occluded instances that are fragmented cannot be detected as a single object. For this 
reason we use this head only for predicting large object candidates which are filtered and refined 
using predictions from other heads.

Due to the fine quantisation of watershed levels, rather than directly predicting the upsampled 
resolution, we gradually refine the lower resolution feature maps while also merging higher 
resolution features from the backbone semantic segmentation network. $F_{hgh}$ is first transformed 
into 
$F_{wtr}$ followed by further refining into $F_W$ as detailed in Table~\ref{tab:BBFNet}. Features 
from the shallowest convolution block of ResNet are then concatenated with $F_W$ and further 
refined with two $1\time1$ convolution to predict the four watershed levels.

We use a weighted cross-entropy loss to train this given by:
\begin{equation}\label{eq:wtr}
 L_{wtr} = \sum_{k\in(0,3)}w_k W_k \; \log(p_k^{wtr}),
\end{equation}

\noindent where $W_k$ is the one-hot ground truth for $k^{th}$ watershed level, $p_k^{wtr}$ its 
predicted probability and $w_k$ its weights.

\begin{table}[t!]
  \begin{center}
    \begin{tabular}{|c|c|c|}
      \hline
       \textbf{Input} & \textbf{Blocks} & \textbf{Output} \\
      \hline
      \hline
       FPN &  dc-256-256, dc-256-128  & $F_{ss}$ \\
       $F_{ss}$& ups, cat, conv-512-($C_{\text{\textit{stuff}}}$+$C_{\text{\textit{thing}}}$), ups 
& Segmentation \\
       $F_{ss}$ & 2$\times$dc-128-128 & $F_{hgh}$ \\
       $F_{hgh}$ & ups, cat, conv-512-128, conv-128-(4$\times C_{\text{\textit{thing}}}$), ups & 
Hough\\
       $F_{hgh}$ & 2$\times$dc-128-128 & $F_{wtr}$ \\
       $F_{wtr}$ & ups, cat, conv-512-128, conv-128-16, ups & $F_W^*$ \\
       $F_{wtr}$ & 2$\times$dc-128-128 & $F_{trp}$ \\
       $F_{trp}$ & ups, cat, conv-512-128, conv-128-128, ups & $F_T^*$\\
       \hline
    \end{tabular}
  \end{center}
  \caption{Architecture of BBFNet. $dc$, $conv$, $ups$ and $cat$ stand for 
deformable convolution~\cite{Dai17iccv}, $1\times1$ convolution, upsampling and 
concatenation respectively. The two numbers that follow $dc$ and $conv$ are the input 
and output channels to the blocks.* indicates that more processing is done 
on these blocks as detailed in $\S\ref{sec:watershed}$ and  
$\S\ref{sec:tripletLossNetwork}$.}\label{tab:BBFNet}
\vspace{-0.5cm}
\noindent\hrulefill 
\vspace{-0.5cm}
\end{table}

\subsection{Triplet Loss Network}\label{sec:tripletLossNetwork}
The triplet loss network is used to refine and merge the detected candidate instances in addition to 
detecting new instances. Towards this goal, a popular choice  is to formulate it as an embedding 
problem using triplet loss~\cite{Brabandere17arxiv}. This loss forces features of pixels belonging 
to the same instance to group together while pushing apart features of pixels from different 
instances. Margin-separation loss is usually employed for better instance separation and is given 
by:

\begin{equation}\label{eq:margin}
  L(f_a,f_p,f_n) = \max\big((f_a-f_p)^2 - (f_a-f_n)^2 + \alpha, 0\big),
\end{equation}
\noindent where $f_a$, $f_p$, $f_n$ are the anchor, positive and negative pixel features 
respectively and 
$\alpha$ is the margin. Choosing $\alpha$ is not easy and depends on the complexity of 
the feature space~\cite{Neven19cvpr}. Instead, we opt for a fully-connected network to 
classify the pixel features and formulate it as a binary classification problem:
\begin{equation}\label{eq:trpNetwork}
 T(f_a,f_*) = 
 \begin{cases}
    1              & \text{if, } f_*=f_p, \\
    0              & \text{if, } f_*=f_n,
\end{cases}
\end{equation}
\noindent We use the cross-entropy loss to train this head:
\begin{equation}\label{ed:trpLoss}
 L_{trp} = \sum_{c\in(0,1)} T_c \; \log(p_c^{trp}),
\end{equation}
\noindent $T_c$ is the ground truth one-hot label for the indicator function and $p^{trp}$ 
the predicted probability.

The pixel feature used for this network is a concatenation of $F_T$ (see Table~\ref{tab:BBFNet}), 
its normalised position in the image ($x,y$) and the outputs of the different heads ($p^{seg}$, 
$p^{wtr}$, $\hat{X}_{\text{off}}$, $\hat{Y}_{\text{off}}$, $\sigma_x$ and $\sigma_y$).

\subsection{Training and Inference}\label{sec:training}
We train the whole network along with its heads in an end-to-end fashion using a weighted loss 
function:
\begin{equation}\label{eq:total}
 L_{\text{total}} = \alpha_1 \, L_{ss} + \alpha_2 \, L_{hgh} + \alpha_3 \, L_{wtr} + \alpha_4 \, 
L_{trp}.
\end{equation}

For the triplet loss network, training with all pixels is prohibitively expensive. Instead we 
randomly choose a fixed number of anchor pixels  $N_a$ for each instance. Hard positive examples 
are obtained by sampling from the farthest pixels to the object center and correspond to watershed 
level $k=0$. For hard negative examples, neighbouring instances' pixels closest to the anchor and 
belonging to the same class are given higher weight. Only half of the anchors use hard example 
mining while the rest use random sampling.

We observe that large objects are easily detected by the watershed head while Hough voting based 
center prediction does well when objects are of the same scale. To exploit this observation, we 
detect large object candidates ($I_{L'}$) using connected components on the watershed predictions 
correspond to $k\geq1$ bins. We then filter out candidates whose predicted Hough center 
($I_{L'}^{\text{center}}$) does not fall within their bounding boxes ($BB_{L'}$). These filtered out 
candidates are fragmented regions of occluded objects or false detections. Using the center pixel of 
the remaining candidates ($I_{L''}$) as anchors points, the triplet loss network refines them over 
the remaining pixels allowing us to detect fragmented regions while also improving their boundary 
predictions.

After the initial watershed step, the unassigned \textit{thing} pixels corresponding to $k=0$ and 
primarily belong to small instances. We use mean-shift clustering with fixed bandwidth ($B$) to 
predict candidate object centers, $I_S^{\text{center}}$. We then back-trace pixels voting for their 
centers to obtain the Hough predictions $I_S$.


Finally, from the remaining unassigned pixels we randomly pick an anchor point and test it with the 
other remaining pixels. We use this as candidates regions 
that are filtered ($I_R$) based on their Hough center predictions, similar to the watershed 
candidates. The final detections are the union of these predictions. We summarise these steps in 
algorithm provided in the supplementary material.

\section{Experiments}\label{sec:results}
In this section we evaluate the performance of BBFNet and present the results we obtain. We first 
describe the datasets and the evaluation metrics used. In $\S\ref{sec:implementation}$ we describe 
the implementation details of our network. $\S\ref{sec:ablation}$ then discusses the performance of 
individual heads and how its combination helps improve the overall accuracies. We presents both the 
qualitative and quantitative results in $\S\ref{sec:experiments}$ and show the flexibility of 
BBFNet 
in $\S\ref{sec:flex}$. We end this section by presenting some of the failure cases in 
$\S\ref{sec:error}$ and comparing them with other MoE+BB based approaches.

\noindent \textbf{Datasets:} The Cityscapes dataset~\cite{Cordts16cvpr} contains driving 
scenes with $2048\times1024$ 
resolution images recorded over various cities in Germany and Switzerland. It consists of $2975$ 
densely annotated images training images and a further $500$ validation images. For 
the panoptic challenge, a total of 19 classes are split into 8 \textit{things} and 11 
\textit{stuff} classes. 

Microsoft COCO~\cite{Lin14eccv} is a large scale object detection and segmentation 
dataset with over $118k$ training
(2017 edition) and $5k$ validation images with varying resolutions. The 
labels consists of 133 classes split into 80 \textit{things} and $53$ \textit{stuff}.\\

\noindent \textbf{Evaluation Metrics:} We benchmark using the Panoptic Quality (PQ) measure which 
was proposed in~\cite{Kirillov19cvpra}. This measure comprises of two terms, Recognition 
Quality (RQ) and Segmentation Quality (SQ), to measure individual performance on recognition and 
segmentation tasks:
\begin{equation}\label{eq:pq}
\text{PQ} = \underbrace{\frac{\sum_{(\text{p,g}) \in \text{TP}} 
\text{IoU(p,g)}}{\left|\text{TP}\right|}}_{\text{SQ}}\underbrace{\frac{\left|\text{TP}\right|}{
\left|\text { TP} \right|+\frac { 1 } {2} \left|\text{FP}\right|+\frac{1}
{2} \left|\text{FN}\right|}}_{\text{RQ}},
\end{equation}

\noindent where, IoU is the intersection-over-union measure, $(\text{p,g})$ are the 
matching predicted and ground-truth regions ($> 0.5$ IoU), TP, FP and FN are 
true-positive, false-positive and false-negative respectively.

\begin{table}[t!]
  \begin{center}
  \begin{tabular}{cc}
    \begin{tabular}[b]{|c|c|c|c|c|c|c|c|}
      \hline
       \textbf{W} & \textbf{H} & \textbf{T} & \textbf{PQ} & \textbf{SQ} & 
\textbf{PQ\textsubscript{s}} & 
\textbf{PQ\textsubscript{m}} & \textbf{PQ\textsubscript{l}}\\
      \hline
      \hline
      \cmark & \xmark & \xmark & 44.4 & 75.7 & 1.3 & 24.1 & 57.9 \\
      \xmark & \cmark & \xmark & 49.7 & 78.8 & 11.6 & 37.4 & 44.5 \\
      \xmark & \xmark & \cmark & 55.9 & 79.4 & 10.4 & 45.5 & 72.0 \\
      \cmark & \cmark & \cmark & \textbf{56.6} & \textbf{80.0} & \textbf{12.6} & \textbf{47.7} &  
\textbf{72.5} \\
      \hline
      \end{tabular} &
      \hspace{1cm}
      \includegraphics[width=.3\columnwidth]{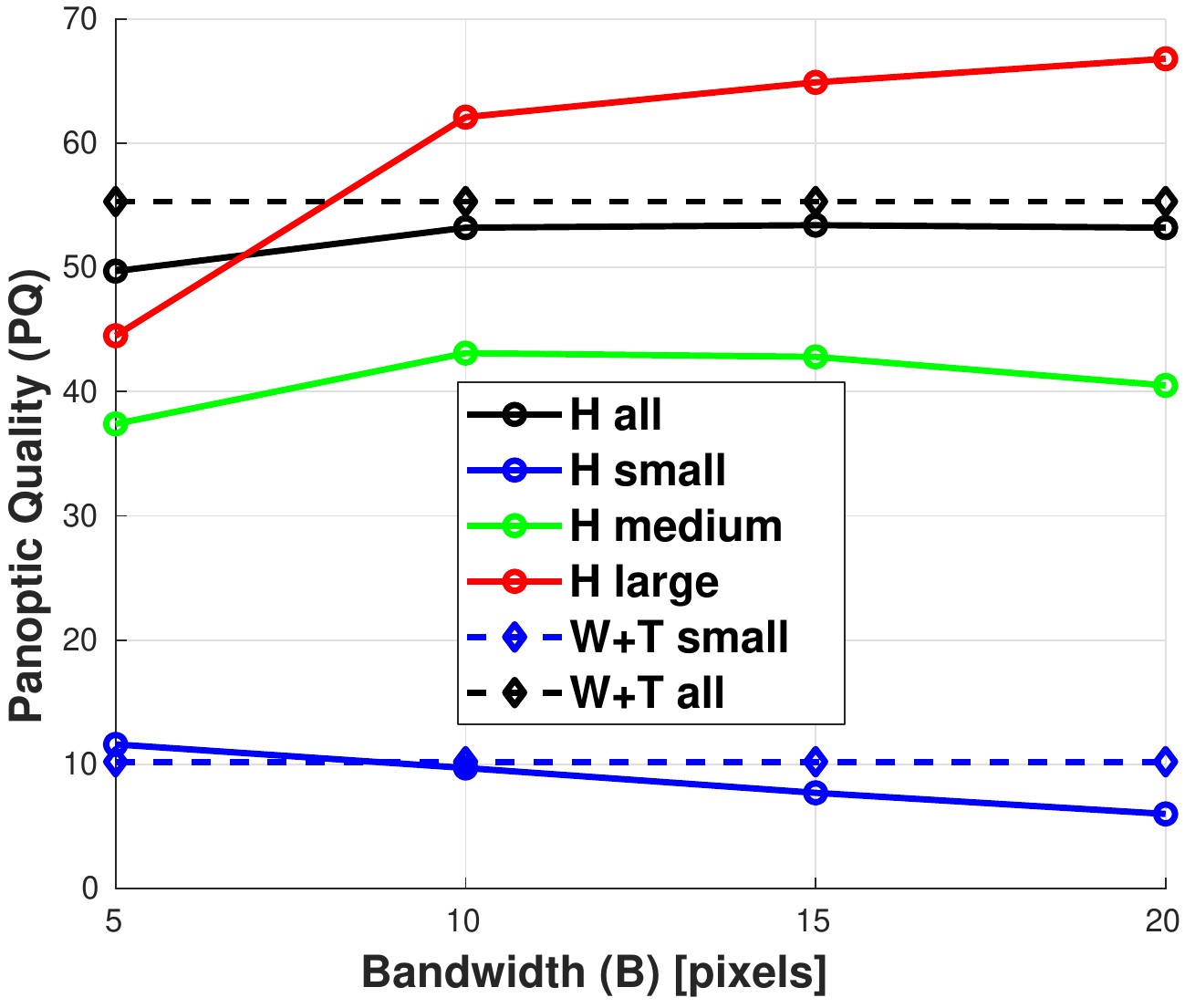} \\
      (a) & \hspace{1cm} (b)\\
      \end{tabular}
  \end{center}
  \caption{(a) Performance of different heads (W- Watershed, H- Hough Voting and T- Triplet Loss 
Network) on Cityscapes validation set. BBFNet exploits the complimentary performance of watershed 
(large objects $>10k$ pixels) and Hough voting head (small objects $<1k$ pixels) resulting in 
higher 
accuracy. PQ\textsubscript{s}, PQ\textsubscript{m} and PQ\textsubscript{l} are the PQ scores for 
small, medium and large objects respectively. Bold is for best results. (b) Performance of Hough 
voting head (H) with varying 
$B$ for different sized objects, $s$-small $<1k$ pixels, $l$-large $>10k$ pixels and $m$-medium 
sized instances. For reference we also plot the performance of Watershed+Triplet loss (W+T) head 
(see Table~\ref{tab:abs}).}\label{tab:abs}
\vspace{-0.5cm}
\noindent\hrulefill 
\vspace{-0.5cm}
\end{table}

\subsection{Implementation Details}\label{sec:implementation}
We use the pretrained ImageNet~\cite{Deng09cvpr} models for ResNet50 and FPN and train the BBFNet 
head from scratch. We keep the backbone fixed for initial epochs before training the whole network 
jointly. In the training loss (\eq~\ref{eq:total}), we set $\alpha_1, \alpha_2, \alpha_3$ and 
$\alpha_4$ parameters to $1.0$, $0.1$, $1.0$ and $0.5$ respectively, since we found this to be a 
good balance between the different losses. The mean-shift bandwidth is set to reduced pixels of 
$B=10$ to help the Hough voting head detect smaller instances. In the watershed head, the number of 
training pixels decreases with $K$ and needs to be offset by higher $w_k$. We found the weights 
$0.2, 0.1, 0.05, 0.01$ to work best for our experiments. Moreover, these weights help the network 
focus on detecting pixels corresponding to lower bins on whom the connected-component is performed. 
To train the triplet-loss network head we set the number of pixels per object $N_a = 1000$. For 
smaller instance, we sample with repetition so as to give equal importance to objects of all sizes. 
All experiments were performed on NVIDIA Titan 1080Ti.

\begin{figure}[t!]
\centering
\includegraphics[width=.45\textwidth]{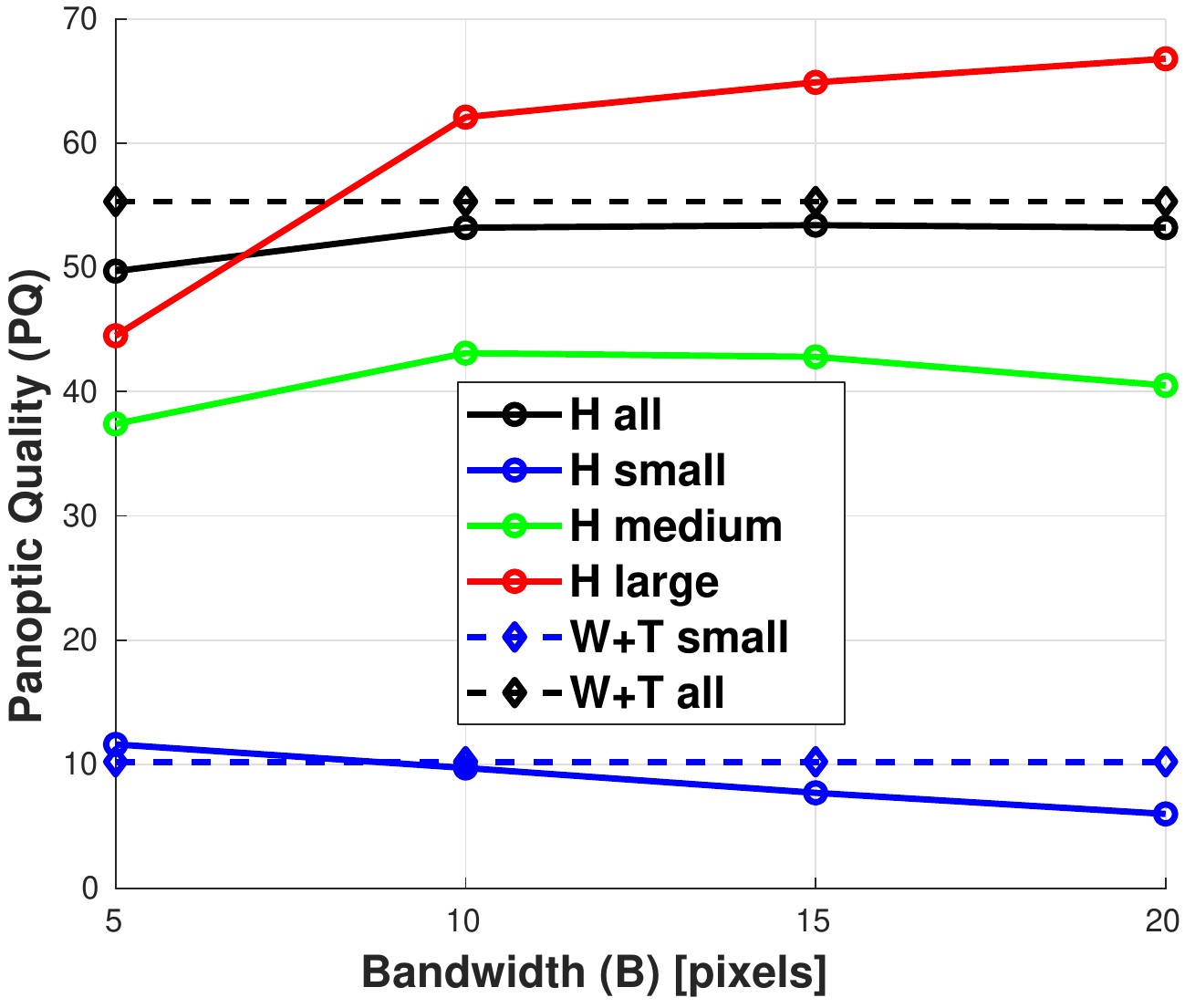}
\caption{Performance of Hough voting head (H) with varying $B$ for different sized objects, 
$s$-small $<1k$ pixels, $l$-large $>10k$ pixels and $m$-medium sized instances. For reference we 
also plot the performance of Watershed+Triplet loss (W+T) head (see 
Table~\ref{tab:abs}).}\label{fig:HghPQ}
\noindent\hrulefill 
\end{figure}

To improve robustness we augment the training data by randomly cropping the images and adding alpha 
noise, flipping and affine transformations. No additional augmentation was used during testing. All 
experiments were performed on NVIDIA Titan 1080Ti. Cityscapes dataset is trained with full 
resolution. For COCO, the longest edge of each image is resized to $1024$ while keeping the aspect 
ratio same.

A common practice during inference is to remove prediction with low detection probability to avoid 
penalising twice (FP and FN)~\cite{Xiong19cvpr}. In BBFNet, these correspond to regions with poor 
segmentation (class or boundary). We use the mean segmentation probability over the predicted 
region as the detection probability and filter regions with low probability ($< 0.65$). 
Furthermore, 
we also observe boundaries shared between multiple objects to be frequently predicted as different 
instances. We filter these by having a threshold ($0.1$) on the IoU between the segmented 
prediction and its corresponding bounding box.

\subsection{Ablation studies}\label{sec:ablation}
We conduct ablation studies here to show the advantage of each individual head and how BBFNet 
exploits them. Table~\ref{tab:abs}(a) shows the results of our experiments on Cityscapes. We use 
the validation sets for all our experiments. We observe that watershed or Hough voting heads alone 
do not perform well. In the case of watershed head this is because performing connected component 
analysis on $k=1$ level (as proposed in~\cite{Bai17cvpr}) leads to poor SQ. Note that performing the 
watershed cut at $k=0$ is also not optimal as this leads to 
multiple instances that share boundaries being grouped into a single detection. By combining the 
Watershed head with a refining step from the triplet loss network we observe over $10$ point 
improvement in accuracy.

On the other hand, the performance of the Hough voting head depends on the bandwidth $B$ that is 
used. Table~\ref{tab:abs}(b) plots its performance with varying $B$. As $B$ increases from $5$ to 
$20$ pixels we observe an initial increase in overall PQ before it saturates. This is because
while the performance increases on large objects ($>10k$ pixels), it reduces on small ($<1k$ 
pixels) and medium sized objects. However, we observe that at lower $B$ it outperforms the 
Watershed+triplet loss head on smaller objects. We exploit this in BBFNet (see 
$\S\ref{sec:training}$) by using the watershed+triplet loss head for larger objects while using 
Hough voting head primarily for smaller objects.

\begin{figure*}[t!]
\centering
  \setlength{\tabcolsep}{2pt}
  \begin{tabular}{ccc}
  \includegraphics[width=0.32\columnwidth, cfbox=blue 1pt 0pt]{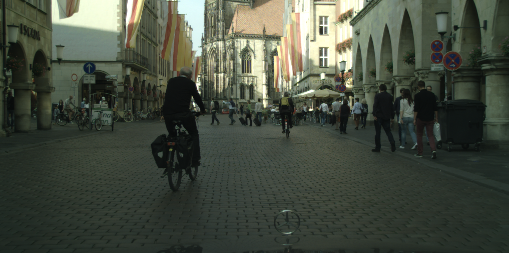}&
  \includegraphics[width=0.32\columnwidth, cfbox=blue 1pt 0pt]{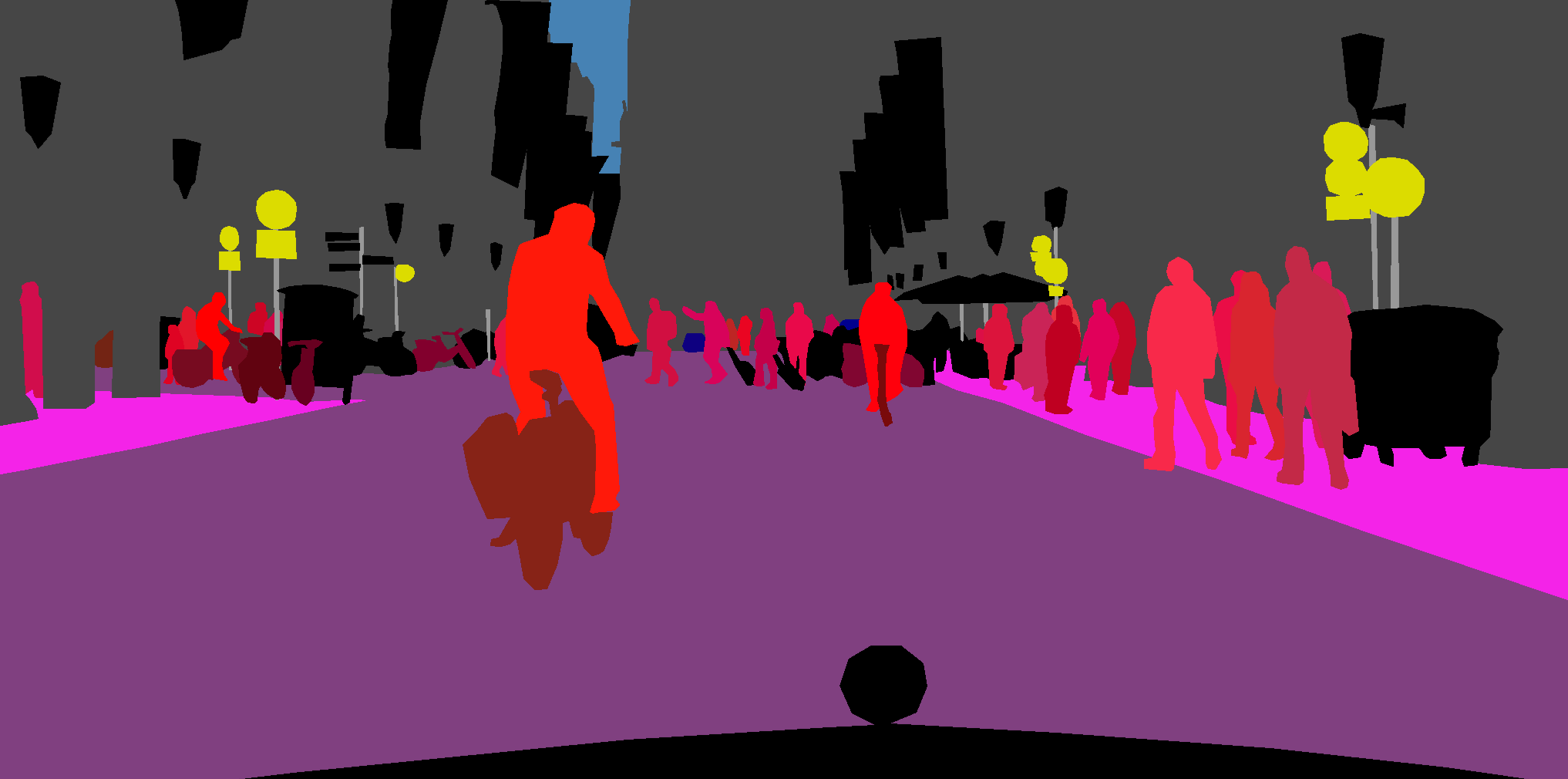}&
  \includegraphics[width=0.32\columnwidth, cfbox=blue 1pt 0pt]{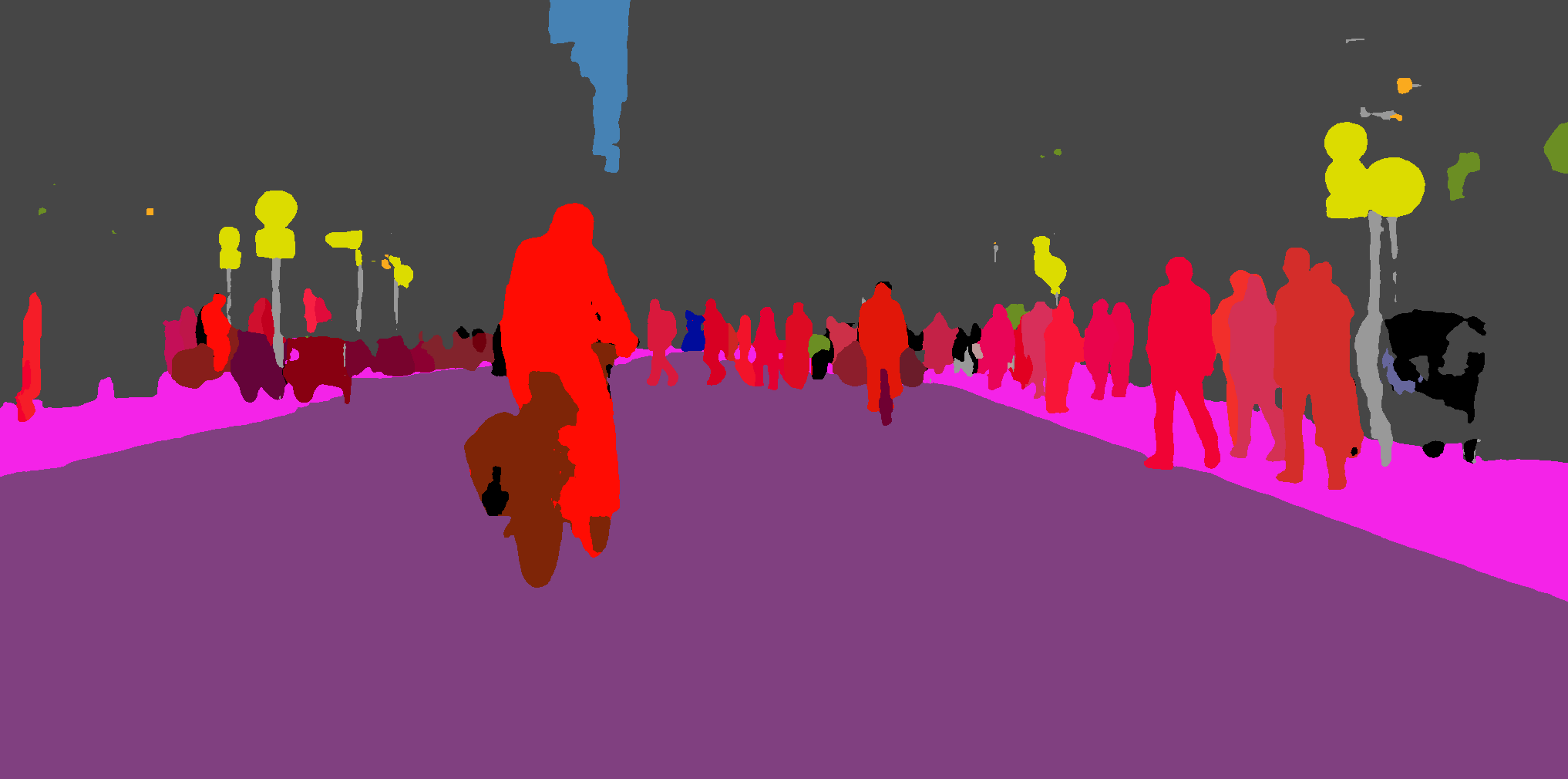}\\
  
  \includegraphics[width=0.32\columnwidth, cfbox=blue 1pt 0pt]{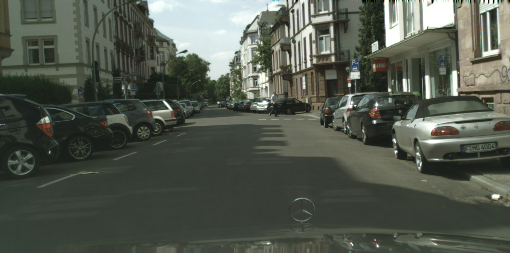}&
  \includegraphics[width=0.32\columnwidth, cfbox=blue 1pt 0pt]{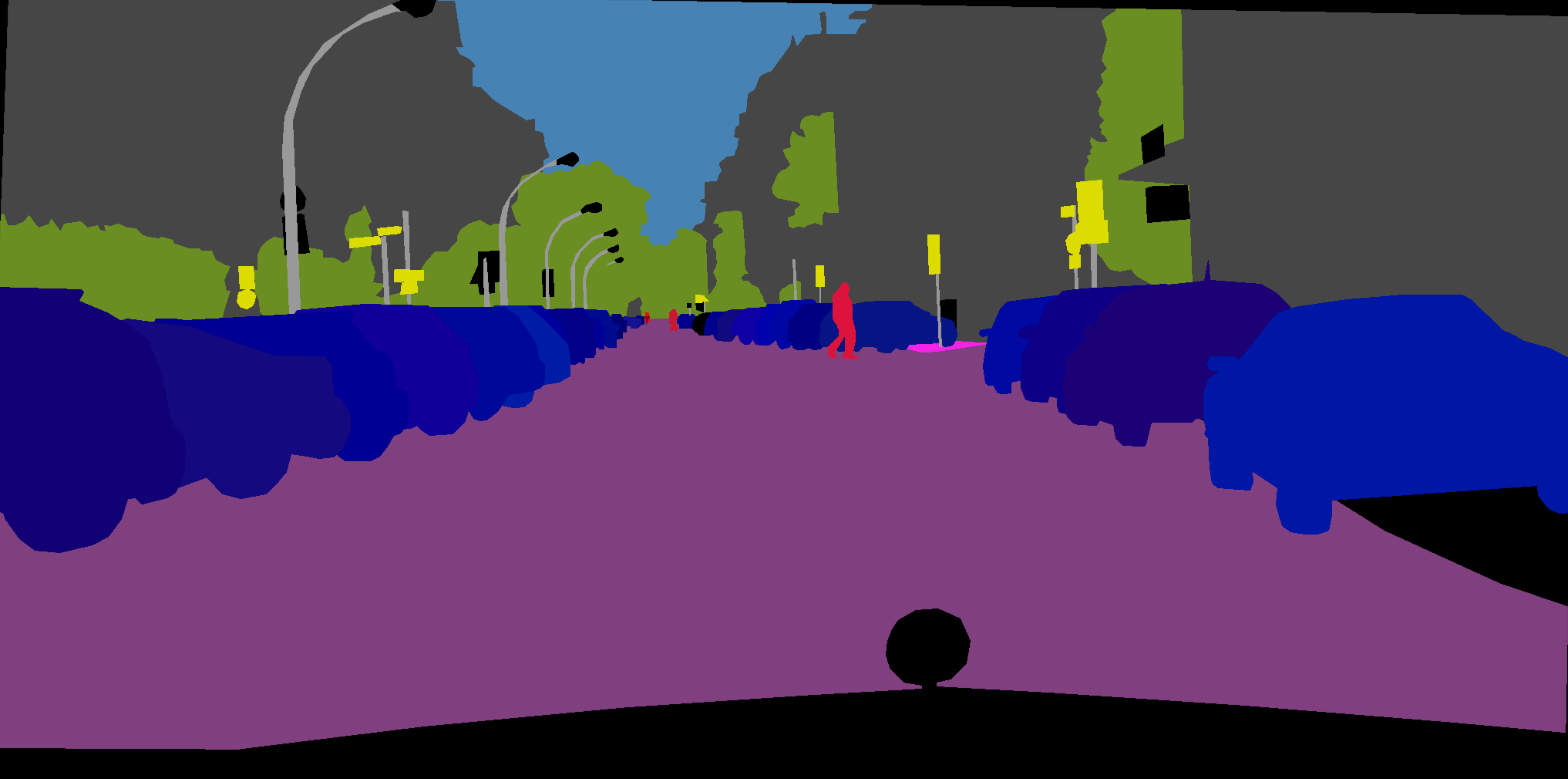}&
  \includegraphics[width=0.32\columnwidth, cfbox=blue 1pt 0pt]{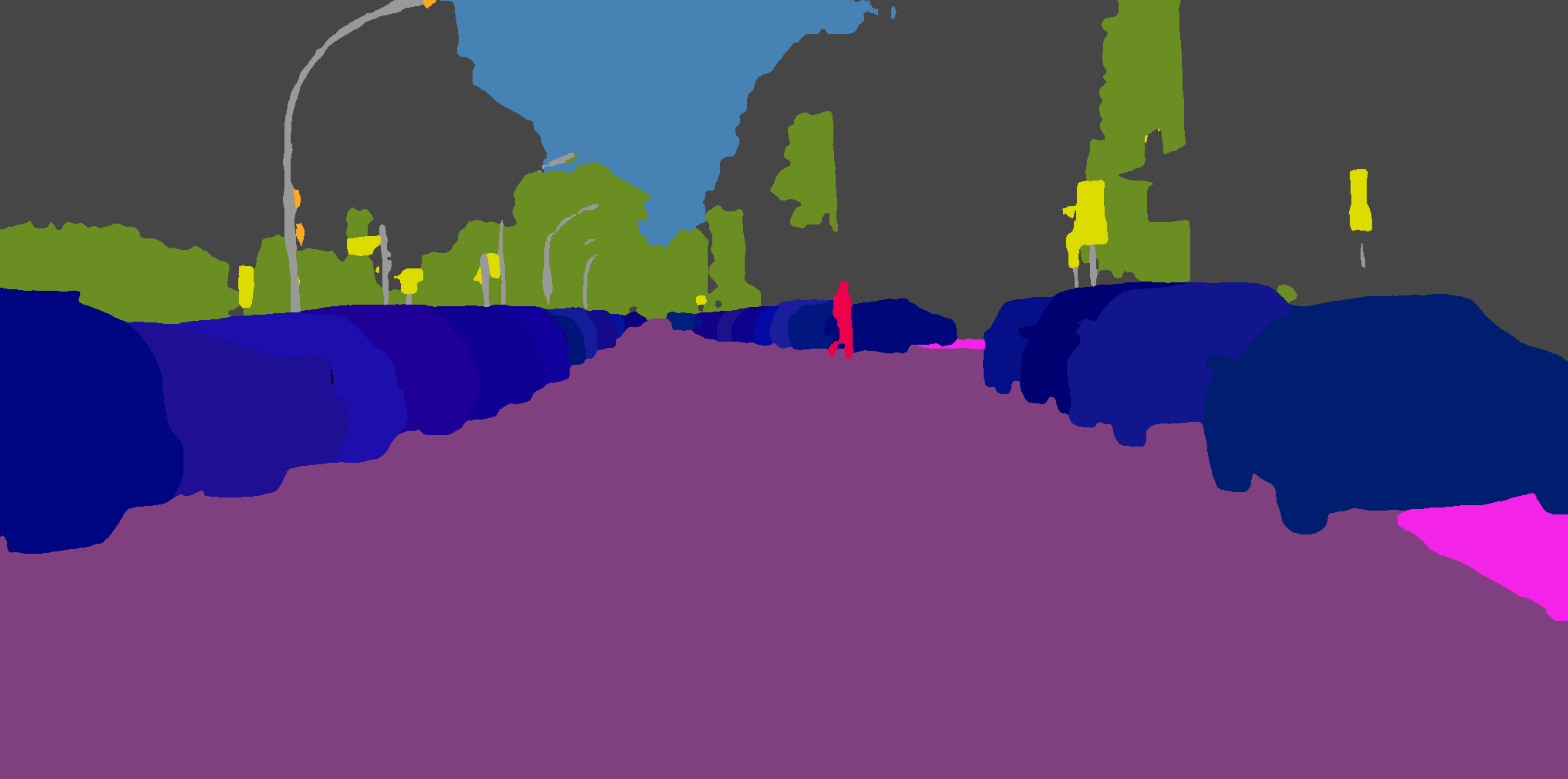}\\
  
  \includegraphics[width=0.32\columnwidth, cfbox=blue 1pt 0pt]{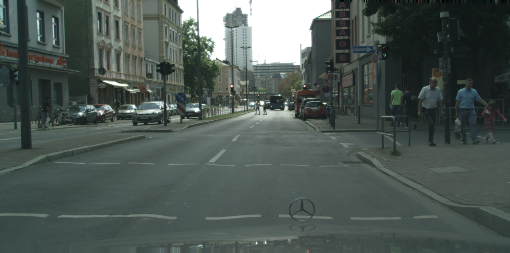}&
  \includegraphics[width=0.32\columnwidth, cfbox=blue 1pt 0pt]{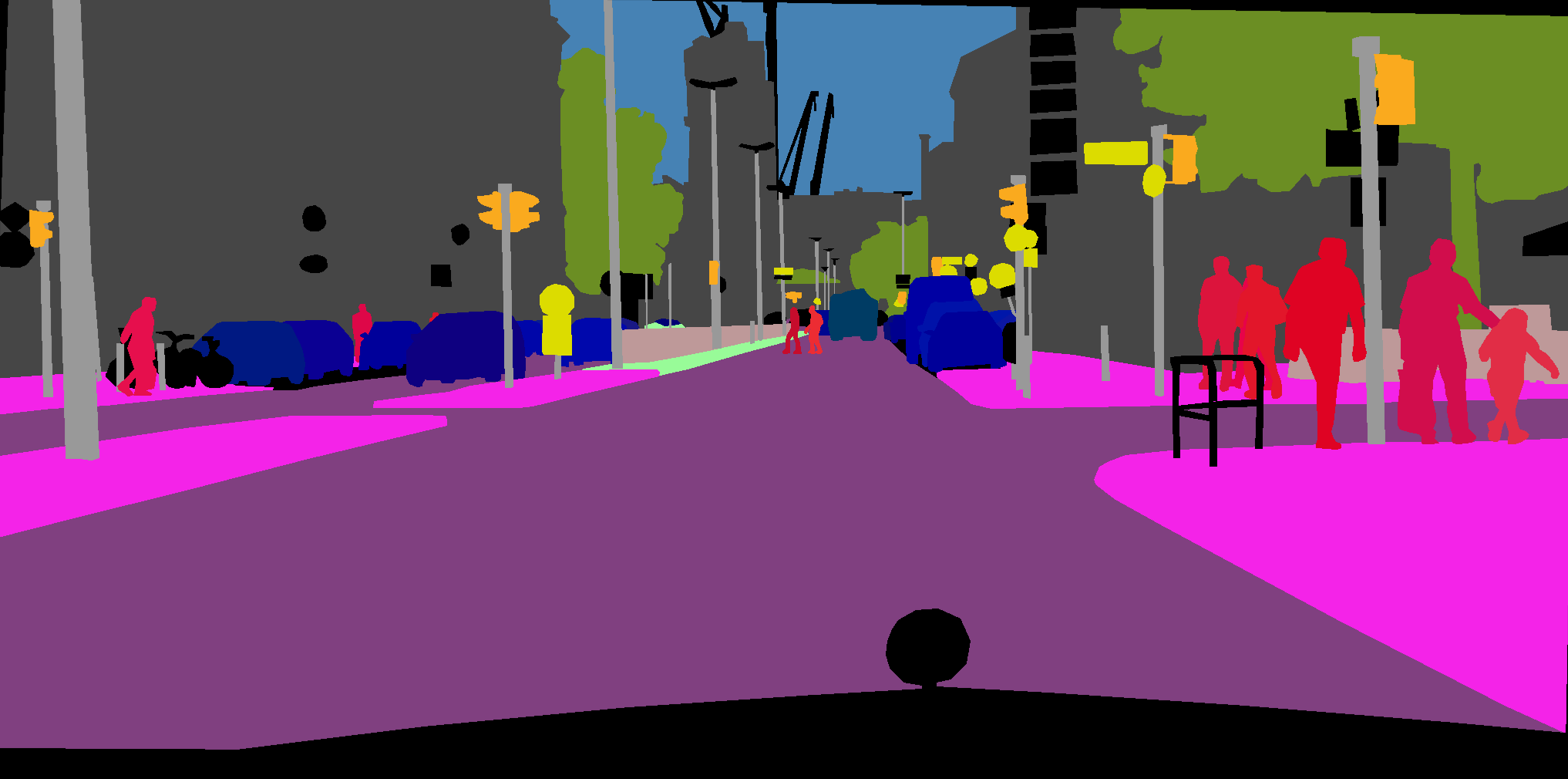}&
  \includegraphics[width=0.32\columnwidth, cfbox=blue 1pt 0pt]{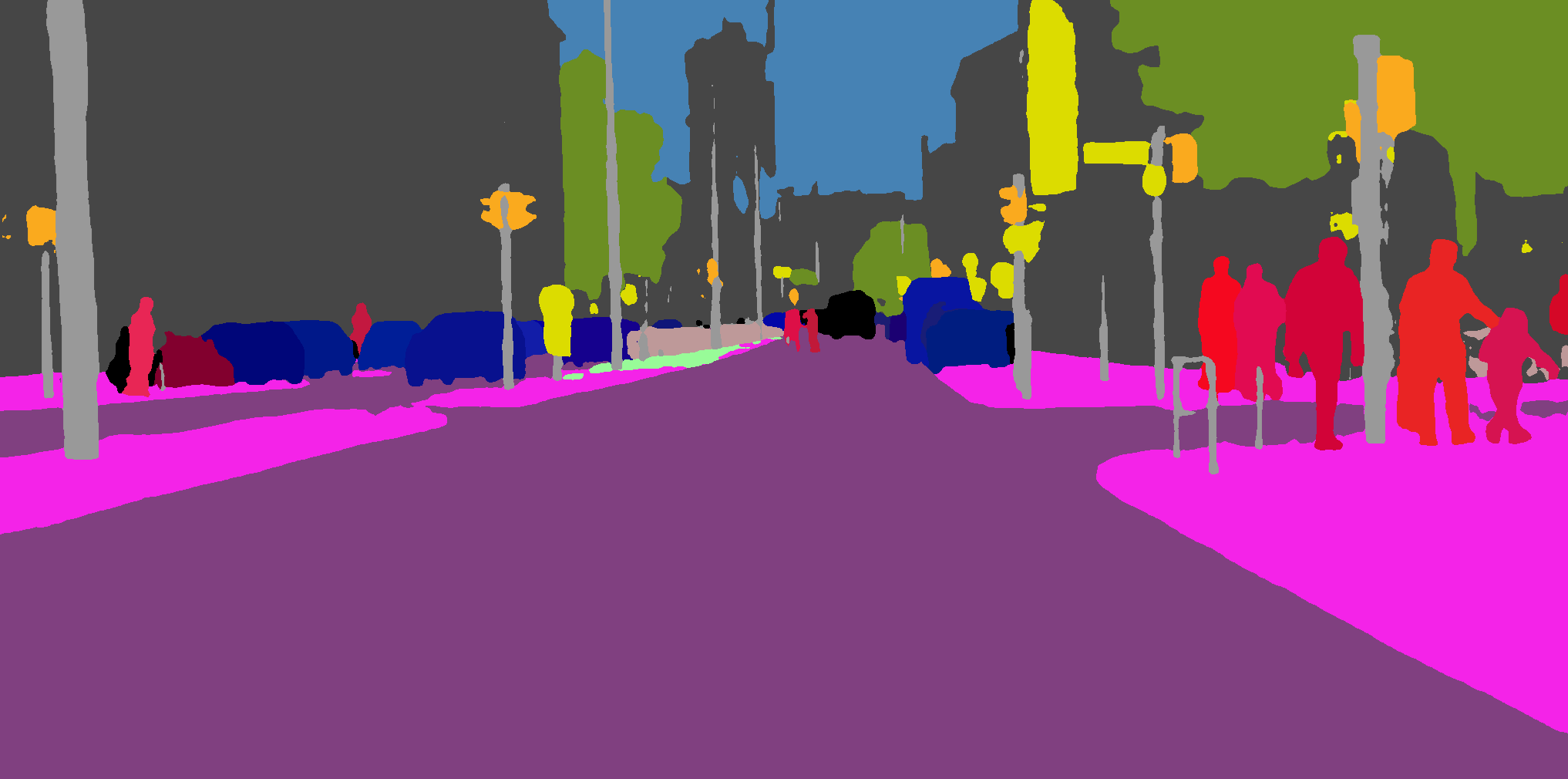}\\
  
  \\
  (a) Input Image & (b) Ground truth &
  (c) BBFNet predictions\\
\end{tabular}
\caption{Sample qualitative results of BBFNet on Cityscapes dataset. BBFNet is able to handle 
complex scenes with multiple occluded and fragmented objects.}\label{fig:city}
\noindent\hrulefill 
\vspace{-0.5cm}
\end{figure*}

\begin{table}[t!]
  \begin{center}
    \begin{tabular}{|c|c||c|c|c|c||c|c|c|c||c|}
      \hline
       \multirow{2}{*}{\textbf{Method}} & \multirow{2}{*}{\textbf{BB}} & 
\multicolumn{4}{c||}{\textbf{Cityscapes}} & 
        \multicolumn{5}{c|}{\textbf{COCO}} \\
       \cline{3-11}
        & & \textbf{PQ} & \textbf{PQ\textsubscript{Th}} & \textbf{PQ\textsubscript{St}} 
        & \textbf{IoU}& \textbf{PQ} & \textbf{PQ\textsubscript{Th}} & \textbf{PQ\textsubscript{St}} 
        & \textbf{IoU} & \textbf{PQ\textsubscript{test}}\\
       \hline
       \hline      
      FPSNet~\cite{Geus19arxiv} & \cmark & 55.1 & 48.3 & 60.1 & - & - & - & - & - & -\\
      Li \etal~\cite{Li20cvpr} & \cmark & \textbf{61.4} & 54.7 & \textbf{66.6} & 77.8 & 
\textbf{43.4} & \textbf{48.6} & \textbf{35.5} & \textbf{53.7} & \textbf{47.2}\textsuperscript{*} \\
      Porzi \etal\cite{Porzi19cvpr} & \cmark & 60.3 & \textbf{56.1} & 63.6 
& 77.5 & - & - & - & - & -\\
      TASCNet~\cite{Li19arxiv} &  \cmark & 55.9 & 50.5 & 59.8 & - & - & - & - & - & 40.7\\
      AUNet~\cite{Li19cvpr} &  \cmark & 56.4 & 52.7 & 59.0 & 73.6 & 39.6 & 49.1 & 25.2 & 45.1 & 
45.2\textsuperscript{*}\\
      P. FPN~\cite{Kirillov19cvpra} & \cmark & 57.7 & 51.6 & 62.2 & 75.0 & 39.0 & 45.9 & 28.7 & 
41.0 & 40.9\textsuperscript{*}\\
      AdaptIS~\cite{Sofiiuk19iccv} & \cmark & 59.0 & 55.8 & 61.3 & 75.3 & 35.9 & 29.3 & 40.3 & - & 
42.8\textsuperscript{*} \\ 
      UPSNet~\cite{Xiong19cvpr} & \cmark & 59.3 & 54.6 & 62.7 & 75.2 & 42.5 & 
48.5 & 33.4 & 54.3 & 46.6\textsuperscript{*}\\
      \hline
      DIN~\cite{Li18eccv} & \xmark & 53.8 & 42.5 & \underline{62.1} & 71.6 & - & - & - & - & -\\
      DeeperLab~\cite{Yang19arxiv} & \xmark & 56.5\textsuperscript{$\pm$} & - & - & - & 
33.8\textsuperscript{$\pm$} & - & - & - & 
34.4\textsuperscript{$\pm$}\\
      SSAP~\cite{Gao19iccv} & \xmark & 56.6 & 49.2 & - & 75.1 & 36.5\textsuperscript{*} 
& - & - & - & 
36.9\textsuperscript{*}\\
      P. DeepLab~\cite{Cheng20cvpr} & \xmark & 
\underline{\textbf{59.7}} & - & - & \underline{\textbf{80.5}} & 
35.1 & - & - & - & 
41.4\textsuperscript{$\pm$} \\
      BBFNet & \xmark & 56.6 & \underline{49.9} & 61.1 & 76.5 & 
\underline{37.1} & \underline{42.9} & \underline{28.5} & \underline{\textbf{54.9}} & 
\underline{42.9}\textsuperscript{*} \\
      \hline
   \end{tabular}
  \end{center}
  \caption{Panoptic segmentation results on the Cityscapes and COCO dataset. All methods use the 
same pretraining (ImageNet) and backbone (ResNet50+FPN), except those with $*$ (ResNet101) and 
$\pm$ (Xception-71). Bold is for overall best results and underscore is the best result in non-BB 
based methods.}\label{tab:cityscapes}
\vspace{-0.5cm}
\noindent\hrulefill 
\vspace{-0.4cm}
\end{table}

\subsection{Experimental Results}\label{sec:experiments}
Table~\ref{tab:cityscapes} benchmarks the performance of BBFNet with existing methods on the 
Cityscapes and COCO datasets. As all 
state-of-the-art methods report results with ResNet50+FPN networks while using the same pre-training 
dataset (ImageNet) we also follow this convention and report our results with this setup except 
where highlighted. 
Multi-scale testing along with horizontal-flipping were used in some works but we omit those results 
here as this can be applied to any existing work including BBFNet to improve performance. From 
the results we observe that BBFNet, without using an MoE or BB, has comparable performance to other 
MoE+BB based methods while outperforming non-BB based methods on the more complicated COCO dataset. 
Fig.~\ref{fig:city} shows some qualitative results on the Cityscapes validation dataset.

\begin{figure*}[t!]
\centering
  \setlength{\tabcolsep}{0.5pt}
  \begin{tabular}{cccccc}
  
  \includegraphics[width=0.245\columnwidth, cfbox=blue 1pt 0pt]{figs/figure6/000000001000.jpg}&
  \includegraphics[width=0.245\columnwidth, cfbox=blue 1pt 0pt]{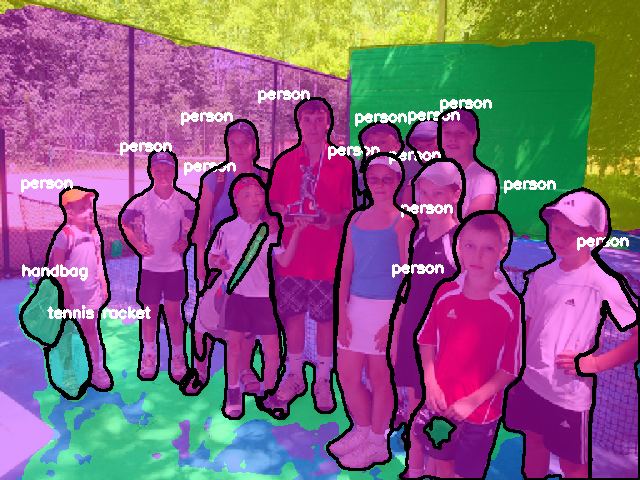}& & &
  \includegraphics[width=0.245\columnwidth, cfbox=blue 1pt 0pt]{figs/figure6/000000166287.jpg}&
  \includegraphics[width=0.245\columnwidth, cfbox=blue 1pt 0pt]{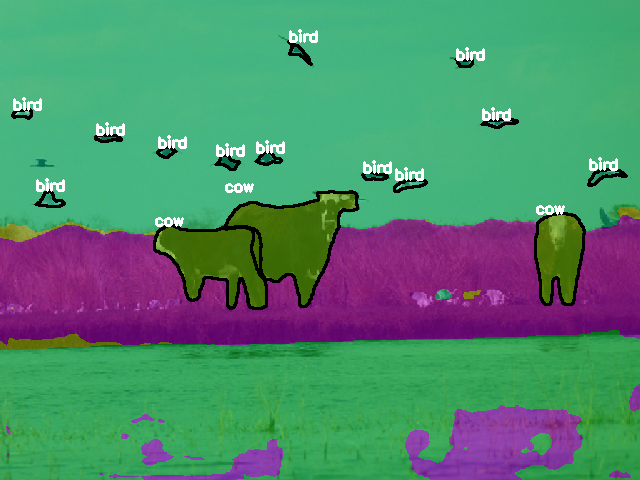}\\
  
  \includegraphics[width=0.245\columnwidth, cfbox=blue 1pt 0pt]{figs/figure6/000000318455.jpg}&
  \includegraphics[width=0.245\columnwidth, cfbox=blue 1pt 0pt]{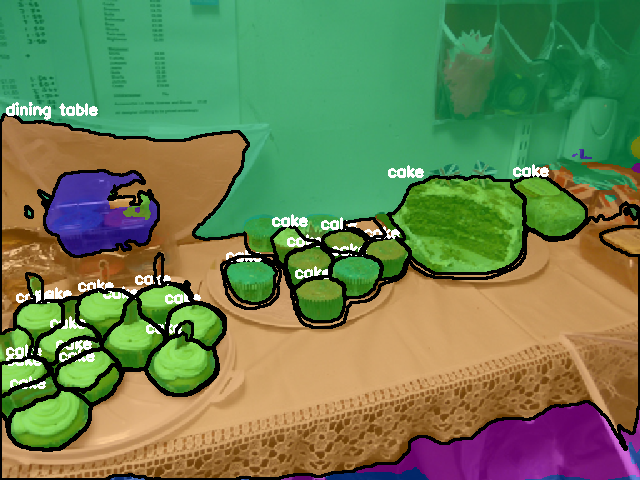}& & &
  \includegraphics[width=0.245\columnwidth, cfbox=blue 1pt 0pt]{figs/figure6/000000127517.jpg}&
  \includegraphics[width=0.245\columnwidth, cfbox=blue 1pt 0pt]{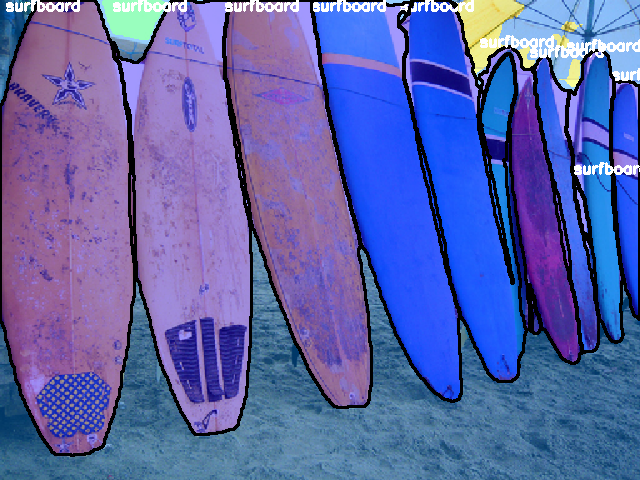}\\
  
  \includegraphics[width=0.245\columnwidth, cfbox=blue 1pt 0pt]{figs/figure6/000000006471.jpg}&
  \includegraphics[width=0.245\columnwidth, cfbox=blue 1pt 0pt]{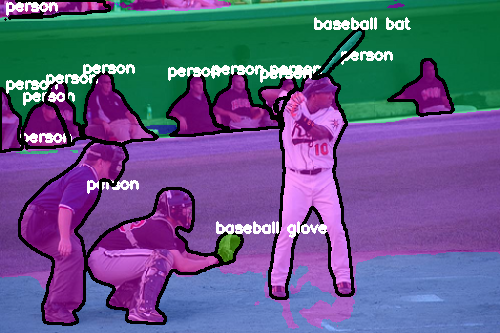}& & &
  \includegraphics[width=0.245\columnwidth, cfbox=blue 1pt 0pt]{figs/figure6/000000152120.jpg}&
  \includegraphics[width=0.245\columnwidth, cfbox=blue 1pt 0pt]{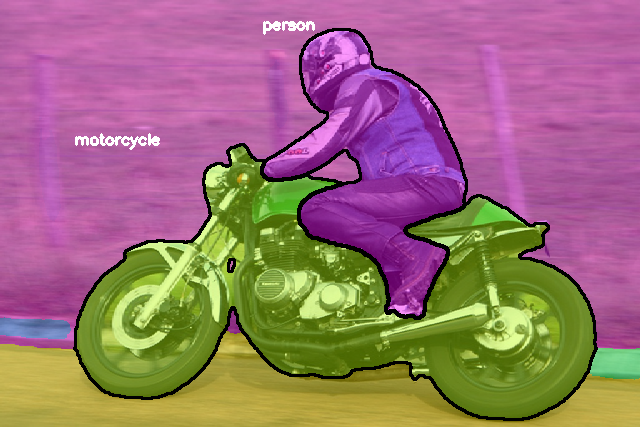}\\
  
  \includegraphics[width=0.245\columnwidth, cfbox=blue 1pt 0pt]{figs/figure6/000000140583.jpg}&
  \includegraphics[width=0.245\columnwidth, cfbox=blue 1pt 0pt]{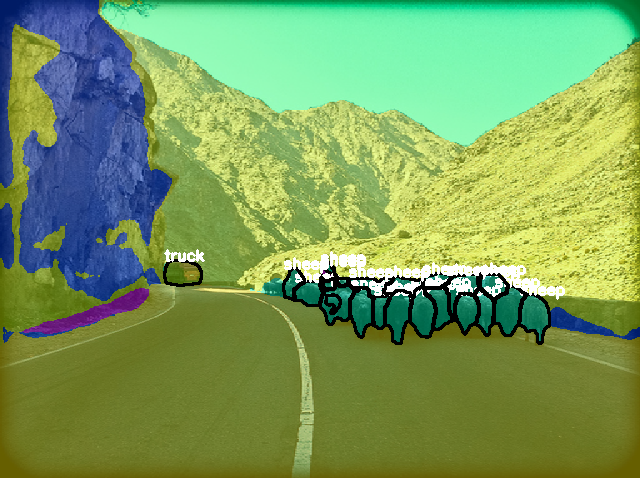}& & &
  \includegraphics[width=0.245\columnwidth, cfbox=blue 1pt 0pt]{figs/figure6/000000131138.jpg}&
  \includegraphics[width=0.245\columnwidth, cfbox=blue 1pt 0pt]{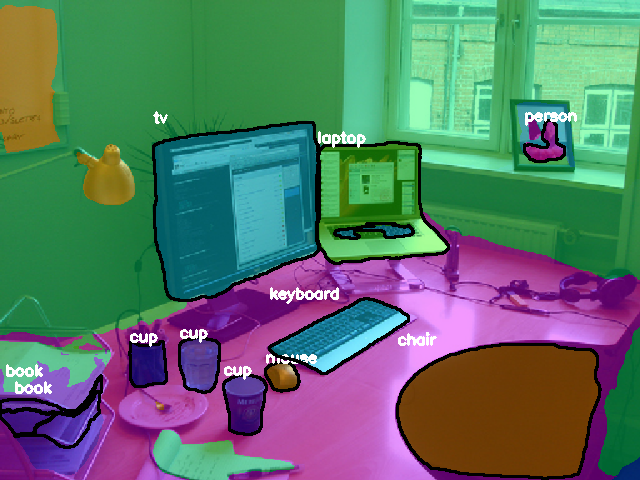}\\
  
  
  
\end{tabular}
\caption{Sample qualitative results of BBFNet on COCO dataset. 
BBFNet can handle different object classes with multiple instances.}\label{fig:coco}
\noindent\hrulefill 
\end{figure*}

\subsection{Flexibility and Efficiency}\label{sec:flex}
To highlight BBFNets ability to work with different segmentation backbones we compare its 
generalisation with different segmentation networks. As it is expected we observe an increase in 
performance with more complex backbones and with DC's but at a cost of reduced efficiency (see 
Table~\ref{tab:generalization}). For reference we also show the performance of a baseline 
proposal-based approach (UPSNet) and a proposal-free approach (SSAP). We used the author provided 
code of UPSNet\footnote{Source code available from \url{https://github.com/uber-research/UPSNet}} 
for computing efficiency figures. Note, that since UPSNet uses Mask R-CNN its backbone cannot be 
replaced and it is not as flexible as BBFNet.

\begin{table}[b!]
  \begin{center}
    \begin{tabular}{|c|c||c|c|c|c||c|c|c|c|}
      \hline
       \multirow{2}{*}{\textbf{Network}} & \multirow{2}{*}{\textbf{Backbone}} & 
\multirow{2}{*}{\textbf{PQ}} & \multirow{2}{*}{\textbf{SQ}} & \multirow{2}{*}{\textbf{RQ}} & 
\multirow{2}{*}{\textbf{IoU}} & \textbf{Flops} & \textbf{Params} & \textbf{T}\textsubscript{I} & 
\textbf{T}\textsubscript{PP} \\
        &  &  &  &  &  & (T) & (M) & (sec) & (sec) \\
       \hline
       \hline
      \multirow{4}{*}{BBFNet} & ERFNet~\cite{Romera18its} (w/o DC) & 46.8 & 76.2 & 58.7 & 69.0 & 
\textbf{0.07} & 
\textbf{2.58} & \textbf{0.12} & \textbf{0.15}\\
       & MobileNetV2~\cite{Sandler18cvpr} & 48.2 & 77.0 & 60.3 & 70.1 & 0.1 & 4.26 & 0.22 & 
0.15\\
       & ResNet50~\cite{He15cvpr} (w/o DC) & 50.4 & 76.3 & 62.4 & 68.9 & 0.49 & 28.2 & 
0.16 & 0.15\\
       & ResNet50~\cite{He15cvpr} & 56.6 & 80.0 & 69.3 & 76.5 & 0.38 & 29.5 & 0.32 & 0.15\\
       & ResNet101~\cite{He15cvpr} & 57.8 & \textbf{80.7} & 70.2 & 
\textbf{78.6} & 0.53 & 48.49 & 0.34 & 0.15 \\ 
      \hline
      \hline
      SSAP~\cite{Gao19iccv} & ResNet50~\cite{He15cvpr} & 56.6 & - & - & 75.1 & - & - & - & 
$\ge$0.26\\
      UPSNet~\cite{Xiong19cvpr} & ResNet50~\cite{He15cvpr} & \textbf{59.3} & 79.7 & \textbf{73.0} & 
75.2 & 0.425 & 44.5 & 0.2 & 0.33\\
      \hline
   \end{tabular}
  \end{center}
  \caption{Panoptic segmentation results showing the trade-off between performance and efficiency 
with different semantic segmentation backbones on the Cityscapes dataset. For efficiency we use 
flops, parameters, inference time (T\textsubscript{I}) and the post-processing time 
(T\textsubscript{PP}). We compare this with a baseline proposal based network (UPSNet) and a 
proposal free network (SSAP).}\label{tab:generalization}
\vspace{-0.5cm}
\noindent\hrulefill 
\vspace{-0.5cm}
\end{table}

As BBFNet does not use a separate instance segmentation head, it is computationally more efficient 
using only $\approx 28.6M$ parameters compared to $44.5M$ UPSNet. We find a similar pattern when we 
compare the number of FLOPs on a $1024\times2048$ image with BBFNet taking $0.38$ TFLOPs compared 
to $0.425$ TFLOPs of UPSNet when using the same ResNet50 backbone. The authors of 
SSAP~\cite{Gao19iccv} do not provide details about their number of parameters, FLOPs and inference 
time. However, they provide timing information for their post-processing step which 
is a cascaded graph partitioning approach that uses the predictions of a semantic segmentation 
network and an affinity pyramid network. This cascaded graph partition module solves a multicut 
optimisation problem~\cite{Keuper15iccv} and takes between $0.26-1.26$ seconds depending on the 
initial resolution for the cascaded graph partition. We believe that BBFNet post-processing step is 
simpler and presumably faster than the one in SSAP.
\begin{figure*}[t!]
\centering
  \setlength{\tabcolsep}{0.5pt}
  \tiny{
  \begin{tabular}{cccc}
  \includegraphics[width=0.245\columnwidth, cfbox=blue 1pt 0pt]{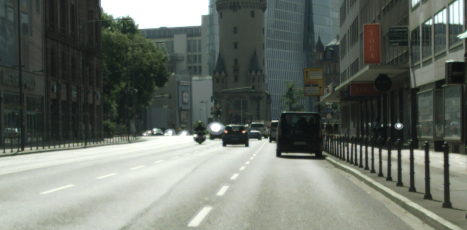}&
  \includegraphics[width=0.245\columnwidth, cfbox=blue 1pt 0pt]{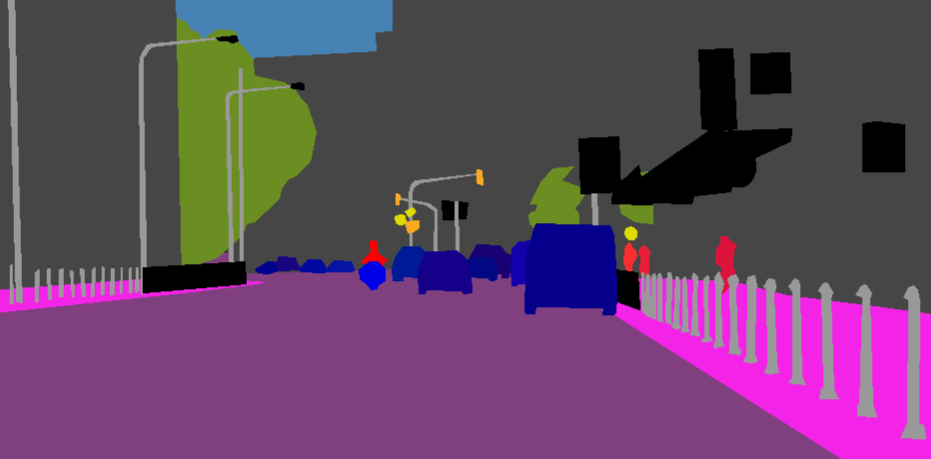}&
  \includegraphics[width=0.245\columnwidth, cfbox=blue 1pt 0pt]{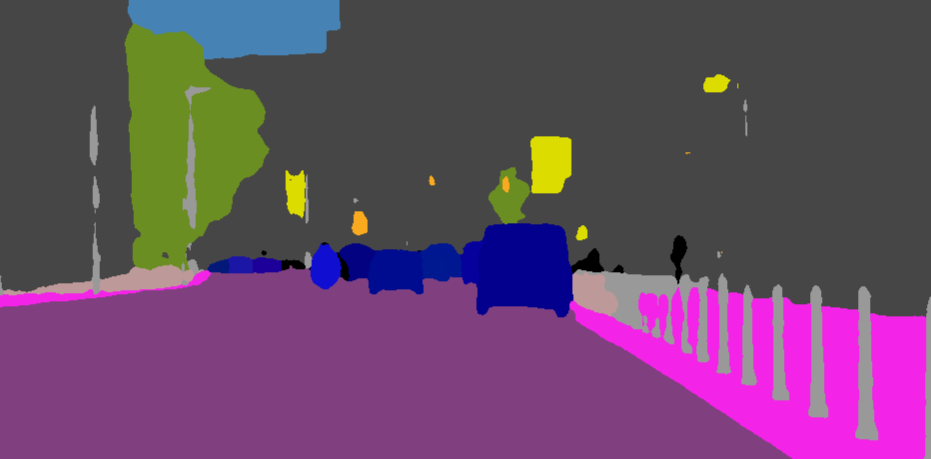}&
  \includegraphics[width=0.245\columnwidth, cfbox=blue 1pt 0pt]{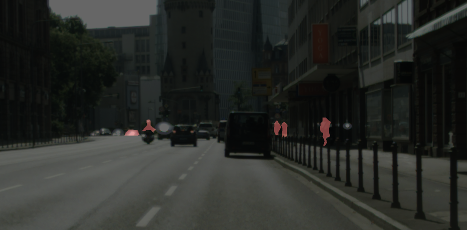}\\
  
  \includegraphics[width=0.245\columnwidth, cfbox=blue 1pt 0pt]{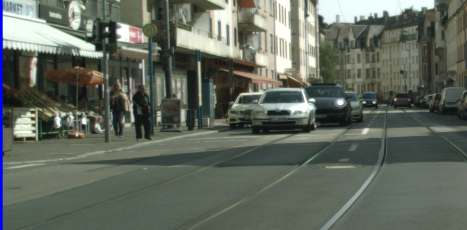}&
  \includegraphics[width=0.245\columnwidth, cfbox=blue 1pt 0pt]{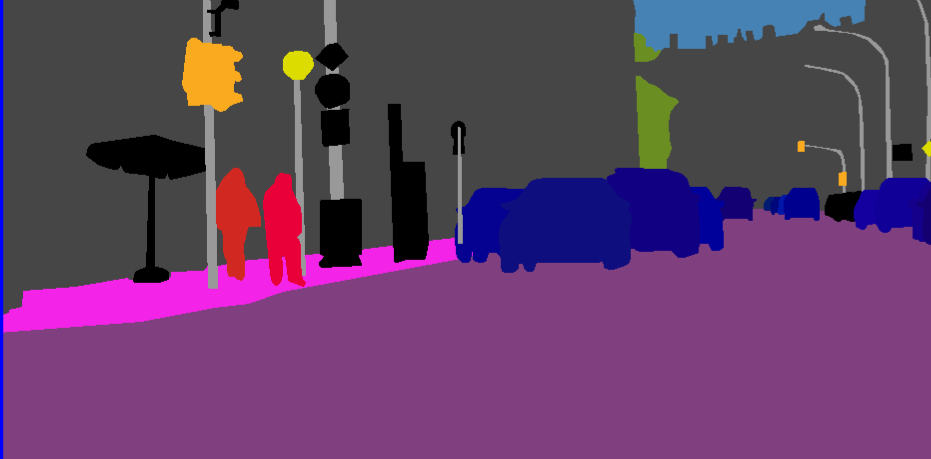}&
  \includegraphics[width=0.245\columnwidth, cfbox=blue 1pt 0pt]{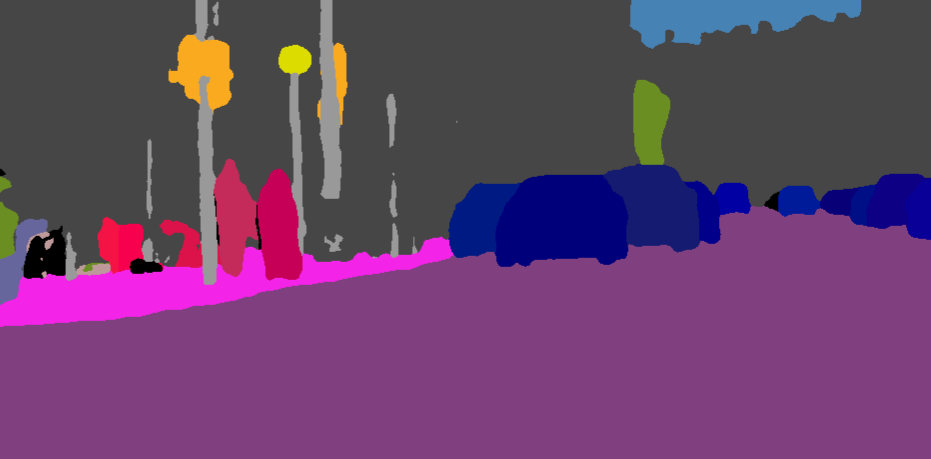}&
  \includegraphics[width=0.245\columnwidth, cfbox=blue 1pt 0pt]{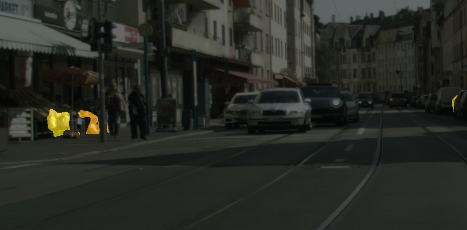}\\
  \\
  (a) Input Image & (b) Ground truth &
  (c) BBFNet predictions & (d) Incorrect predictions\\
\end{tabular}}
\caption{Sample results where BBFNet fails. First row shows an example where low 
confidence of semantic segmentation network leads to missed detection while the second row shows 
examples of false positives due to wrong class label prediction. Without MoE these errors 
from the semantic segmentation network cannot be corrected by BBFNet.}\label{fig:errors}
\noindent\hrulefill 
\vspace{-0.5cm}
\end{figure*}

\begin{table}[b!]
\vspace{-0.5cm}
  \begin{center}
    \begin{tabular}{|c||c|c|c||c|c|c||c|c|c|}
      \hline
       \multirow{2}{*}{\textbf{Network}} & \multicolumn{3}{c||}{\textbf{Small}} &  
\multicolumn{3}{c||}{\textbf{Medium}} &  
\multicolumn{3}{c|}{\textbf{Large}}\\
      \cline{2-10}
      & \textbf{TP} & \textbf{FP} & \textbf{FN} & \textbf{TP} & \textbf{FP} & \textbf{FN} & 
\textbf{TP} & \textbf{FP} & \textbf{FN} \\
      \hline 
      \hline
      UPSNet & 1569 & 722 & 2479 & 3496 & 401 & 954 & 1539 & 49 & 82 \\
      BBFNet & 1067 & 666 & 2981 & 3446 & 680 & 1004 & 1527 & 82 & 94 \\
      \hline
   \end{tabular}
  \end{center}
  \caption{Performance comparison of BBFNet with an MoE+BB method (UPSNet). Due to a 
non-MoE approach, errors from the backbone semantic segmentation network (low TP-small and 
high FP-medium, large) cannot be corrected by BBFNet.}\label{tab:errors}
\vspace{-0.5cm}
\noindent\hrulefill 
\vspace{-0.4cm}
\end{table}

\subsection{Error Analysis}\label{sec:error}
We discuss the reasons for performance difference between our bounding-box free method and ones 
that use bounding-box proposals. UPSNet~\cite{Xiong19cvpr} is used as a benchmark as it shares 
common features with other methods. Table~\ref{tab:errors} depicts the number of predictions made 
for different sized objects in the Cityscapes validation dataset. We report the True Positive (TP), 
False Positive (FP) and the False Negative (FN) values.

One of the areas where BBFNet performs poorly is the number of small object detections. BBFNet 
detects $2/3$ of the smaller objects compared to UPSNet. Poor segmentation (wrong class label 
or inaccurate boundary prediction) also leads to a relatively higher FP for medium and large sized 
objects. Figure~\ref{fig:errors} shows some sample examples. The multi-head MoE approach helps 
addressing these issues but at the cost of additional complexity and computation 
time ($\S\ref{sec:experiments}$). For applications where time or memory are more critical 
compared to detecting smaller objects, BBFNet would be a more suited solution.

\section{Conclusions and Future Work}\label{sec:conclusions}
We presented an efficient bounding-box free panoptic segmentation method called BBFNet. Unlike 
previous methods, BBFNet does not use any instance segmentation network to predict \textit{things}. 
It instead refines the boundaries from the semantic segmentation output obtained from any 
off-the-shelf segmentation network. This allows us to be flexible while out-performing 
proposal-free methods on the more complicated COCO benchmark.

In the next future we would work on making the network end-to-end trainable and improving the 
efficiency by removing the use of DCN while maintaining similar accuracy.

\section{Acknowledgment}
We would like to thank Prof. Andrew Davison and Dr. Alexandre Morgand for their critical feedback 
during the course of this work.

\clearpage
{\small
\bibliographystyle{splncs03}

}
\newpage
\section {Supplementary}\label{sec:supplementary}
\subsection{Inference Algorithm}
We summarise the inference steps detailed in $\S\ref{sec:training}$ in an 
algorithm~\ref{alg:inference}
\begin{algorithm}[t]
 \caption{Compute Instance segments $I$}\label{alg:inference}
 \scriptsize
 \begin{algorithmic}[1]
  \Require Watershed levels $W_k$, predicted class $c$, probability $p^{ss}_c$, 
$\hat{X}_{\text{center}}$ and 
$\hat{Y}_{\text{center}} \vee c \in C_{\text{\textit{thing}}}$
  \State $I_{L'} \leftarrow$ connected-components on $W_k >= 1$. \Comment{Large instance
candidates} 
  \State $BB_{I_{L'}} \leftarrow$ bounding-box of $I_{L'}$. 
  
  \State $I_{L'}^{\text{center}} \leftarrow \frac{\sum_{x\in I_{L'}} \hat{X}_{\text{center}} 
p^{ss}_c}{\sum_{x\in I_{L'}}
p^{ss}_c}$, $\frac{\sum_{y\in I_{L'}} \hat{Y}_{\text{center}} p^{ss}_c}{\sum_{y\in I_{L'}} 
p^{ss}_c}$.

  \State $I_{L''} \leftarrow I_{L'}^{\text{center}} \in BB_{I_{L'}}$.  \Comment{Filter
candidates} 
    
  \State $I_L \leftarrow I_{L''} \cup T(f_a,f_*) = 1 \vee$ $f_a=I_{L''}^{\text{center}}$ \& 
$f_*= c \in C_{\text{\textit{thing}}}/I_{L''}$.~\ref{eq:trpNetwork} 

  \State $I_S^{\text{center}} \leftarrow$ meanshift $\vee c \in C_{\text{\textit{thing}}}/I_{L}$. 
\Comment{Small instances} 
  
  \State $I_S \leftarrow$ Back-trace pixels voting for $I_S^{\text{center}}$

  \While {$c \notin \emptyset$} \Comment{Remaining instances}  
  \State $I_{R'} \leftarrow (\cup T(f_a,f_*) = 1) \vee$ $f_a=$ Random(c) \& 
$f_*=c \in C_{\text{\textit{thing}}}/I_{L}/I_S$.~\ref{eq:trpNetwork} 
 \EndWhile
 \State $BB_{I_{R'}} \leftarrow$ bounding-box of $I_{R'}$. 
 \State $I_{R'}^{\text{center}} \leftarrow \frac{\sum_{x\in I_{R'}} \hat{X}_{\text{center}} 
p^{ss}_c}{\sum_{x\in I_{R'}}
p^{ss}_c}$, $\frac{\sum_{y\in I_{R'}} \hat{Y}_{\text{center}} p^{ss}_c}{\sum_{y\in I_{R'}} 
p^{ss}_c}$.

  \State $I_R \leftarrow I_{R'}^{\text{center}} \in BB_{I_{R'}}$.  \Comment{Filter
candidates} 
 
 \State $I \leftarrow I_L \cup I_S \cup I_R$
 \end{algorithmic}
\end{algorithm}

\subsection {Cityscapes dataset}
Table~\ref{tab:city} gives the per-class results for the Cityscapes dataset. The first 
11 classes are \textit{stuff} while the rest 8 are \textit{thing} label.
\begin{table}[htpb]
  \begin{center}
   \begin{tabular}{|c|c|c|c|c|c|c|}
    \hline
    \textbf{class} & \textbf{PQ} & \textbf{SQ} & \textbf{RQ} & 
    \textbf{PQ$_s$} & \textbf{PQ$_m$} & \textbf{PQ$_l$}\\
    \hline
    \hline
    road & 97.9 & 98.2 & 99.7 & 0.0 & 0.0 & 0.0 \\
    sidewalk & 74.9 & 84.0 & 89.2 & 0.0 & 0.0 & 0.0\\
    building & 87.4 & 89.2 & 98.0 & 0.0 & 0.0 &  0.0\\
    wall & 26.2 & 72.0 & 36.4 & 0.0 & 0.0 & 0.0\\
    fence & 27.6 & 72.9 & 37.8 & 0.0 & 0.0 & 0.0\\
    pole & 50.8 & 65.2 & 77.9 & 0.0 & 0.0 & 0.0\\
    T. light & 40.7 & 68.4 & 59.4 & 0.0 & 0.0 & 0.0\\
    T. sign & 64.8 & 76.4 & 84.7 & 0.0 & 0.0 & 0.0\\
    vegetation & 88.3 & 90.3 & 97.8 & 0.0 & 0.0 & 0.0\\
    terrain & 27.6 & 72.4 & 38.1 & 0.0 & 0.0 & 0.0\\
    sky & 85.1 & 91.9 & 92.7 & 0.0 & 0.0 & 0.0\\
    person & 48.0 & 76.3 & 62.9 & 22.9 & 62.0 & 81.9\\
    rider & 43.8 & 71.2 & 61.6 & 11.2 & 54.3 & 71.7\\
    car & 64.7 & 84.5 & 76.5 & 32.2 & 72.2 & 91.5\\
    truck & 48.2 & 84.5 & 57.0 & 6.7 & 37.3 & 72.3\\
    bus & 69.1 & 88.5 & 78.1 & 0.0 & 49.6 & 85.0\\
    train & 46.1 & 80.7 & 57.1 & 0.0 & 10.7 & 64.2\\
    motorcycle & 36.9 & 72.5 & 50.9 & 8.9 & 44.3 & 56.6\\
    bicycle & 40.6 & 70.2 & 57.9 & 17.4 & 47.1 & 56.8 \\
    \hline
  \end{tabular}
  \end{center}
  \caption{Per-class results for cityscapes dataset. The first 11 classes are from \textit{stuff} 
while the rest 8 are from \textit{thing} label.}\label{tab:city}
\end{table}

\subsection {COCO dataset}
Tables~\ref{tab:coco1}, ~\ref{tab:coco2} and ~\ref{tab:coco3} give the per-class results for the 
COCO dataset. The first 80 classes are \textit{things} while the rest 53 are \textit{stuff} label.
\begin{table}[htpb]
  \begin{center}
    \begin{tabular}{|c|c|c|c|c|c|c|}
    \hline
    \textbf{class} & \textbf{PQ} & \textbf{SQ} & \textbf{RQ} & 
    \textbf{PQ$_s$} & \textbf{PQ$_m$} & \textbf{PQ$_l$}\\
    \hline
    \hline
person & 51.7 & 77.7 & 66.5 & 32.0 & 55.7 & 71.1\\
bicycle & 17.6 & 66.9 & 26.4 & 7.9 & 19.5 & 33.2\\
car & 42.1 & 81.0 & 52.0 & 30.9 & 54.9 & 56.0\\
motorcycle &  40.6 & 74.1 & 54.8 & 13.7 & 35.7 & 58.9\\
airplane & 56.8 & 78.0 & 72.7 & 45.4 & 37.5 & 72.3\\
bus & 52.0 & 87.8 & 59.3 & 0.0 & 34.1 & 76.4\\
train & 50.0 & 84.2 & 59.4 & 0.0 & 16.8 & 56.2\\
truck & 24.3 & 78.4 & 31.0 & 13.3 & 21.5 & 36.7\\
boat & 23.1 & 68.2 & 33.9 & 10.9 & 32.2 & 37.5\\
T. light & 36.7 & 77.3 & 47.4 & 31.4 & 51.5 & 69.8\\
F. hydrant & 77.5 & 87.1 & 88.9 & 0.0 & 71.6 & 91.3\\
S. sign & 80.4 & 91.3 & 88.0 & 36.5 & 88.5 & 92.6\\
P. meter & 56.2 & 87.9 & 64.0 & 0.0 & 48.6 & 82.0\\
bench & 17.2 & 67.9 & 25.4 & 11.0 & 23.4 & 13.5\\
bird & 28.2 & 73.5 & 38.4 & 15.0 & 47.5 & 78.6\\
cat & 86.3 & 91.2 & 94.6 & 0.0 & 78.7 & 89.0\\
dog & 69.3 & 86.0 & 80.6 & 0.0 & 58.5 & 82.9\\
horse & 56.5 & 78.7 & 71.8 & 0.0 & 47.6 & 71.5\\
sheep & 49.5 & 79.0 & 62.6 & 23.7 & 59.1 & 80.7\\
cow & 42.3 & 82.5 & 51.4 & 0.0 & 32.4 & 70.1\\
elephant & 63.0 & 83.9 & 75.0 & 0.0 & 37.4 & 71.5\\
bear & 64.5 & 85.0 & 75.9 & 0.0 & 56.2 & 75.8\\
zebra & 74.3 & 88.2 & 84.2 & 0.0 & 71.6 & 81.9\\
giraffe & 73.1 & 82.2 & 88.9 & 0.0 & 77.0 & 72.4\\
backpack & 9.6 & 83.5 & 11.5 & 2.8 & 16.4 & 34.7\\
umbrella & 50.2 & 81.9 & 61.3 & 21.6 & 57.2 & 64.3\\
handbag & 12.8 & 74.6 & 17.2 & 2.8 & 20.4 & 29.7\\
tie & 29.8 & 77.6 & 38.5 & 0.0 & 56.2 & 51.4\\
suitcase & 51.6 & 79.9 & 64.6 & 16.7 & 51.6 & 70.2\\
frisbee & 70.4 & 85.8 & 82.1 & 51.1 & 77.7 & 93.0\\
skis & 4.5 & 71.2 &  6.3 & 0.0 & 12.4 &  0.0 \\
snowboard & 24.2 & 65.3 & 37.0 & 9.5 & 34.3 &  0.0\\
kite & 27.1 & 72.4 & 37.5 & 25.8 & 21.7 & 43.6\\
B. bat & 23.8 & 67.9 & 35.0 & 35.0 &  8.5 &  0.0\\
B. glove & 37.7 & 83.6 & 45.2 & 18.6 & 74.3 &  0.0\\
skateboard & 37.3 & 71.5 & 52.2 & 0.0 & 48.8 & 50.6\\
surfboard & 48.5 & 75.2 & 64.4 & 29.8 & 49.0 & 69.0\\
T. racket &  58.1 & 83.0 & 70.0 & 27.1 & 68.6 & 86.7\\
bottle & 38.6 & 80.7 & 47.8 & 29.5 & 49.4 & 81.8\\
wine glass & 38.7 & 79.3 & 48.8 & 0.0 & 44.4 & 86.1\\
cup & 48.5 & 88.1 & 55.0 & 15.9 & 70.9 & 75.6\\
fork & 8.5 & 63.5 & 13.3 & 7.2 & 10.4 &  0.0\\
knife & 17.7 & 78.7 & 22.5 & 0.0 & 26.3 & 68.2\\
spoon & 20.2 & 76.4 & 26.4 & 0.0 & 36.9 &  0.0\\
bowl & 29.9 & 78.6 & 38.0 & 17.3 & 32.2 & 39.8\\
banana & 16.5 & 76.4 & 21.6 & 4.0 & 22.1 & 35.5\\
apple & 30.4 & 87.5 & 34.8 & 8.0 & 63.6 & 51.3\\
sandwich & 31.8 & 88.4 & 36.0 & 0.0 & 34.2 & 32.9\\
orange & 59.8 & 88.3 & 67.7 & 36.1 & 37.9 & 82.8\\
broccoli & 22.4 & 74.9 & 30.0 & 0.0 & 20.3 & 42.6\\
carrot & 17.3 & 74.2 & 23.3 & 12.4 & 24.1 &  0.0\\
hot dog & 26.5 & 68.6 & 38.6 & 13.7 & 29.6 & 27.5\\
pizza & 44.5 & 83.2 & 53.5 & 12.6 & 37.5 & 54.5\\
donut & 44.5 & 86.5 & 51.4 & 45.2 & 26.2 & 72.2\\

    \hline
    \end{tabular}
  \end{center}
  \caption{Per-class results for COCO dataset. Continued in Table~\ref{tab:coco2}}\label{tab:coco1}
\end{table}

\begin{table}[tpb]
  \begin{center}
    \begin{tabular}{|c|c|c|c|c|c|c|}
    \hline
    \textbf{class} & \textbf{PQ} & \textbf{SQ} & \textbf{RQ} & 
    \textbf{PQ$_s$} & \textbf{PQ$_m$} & \textbf{PQ$_l$}\\
    \hline
    \hline
cake & 49.9 & 90.2 & 55.3 & 0.0 & 31.6 & 62.3\\
chair & 24.0 & 74.3 & 32.3 & 7.4 & 33.6 & 41.5\\
couch & 44.1 & 80.8 & 54.5 & 0.0 & 32.4 & 52.4\\
P. plant & 27.2 & 74.1 & 36.7 & 16.6 & 33.1 & 27.3\\
bed & 48.4 & 82.0 & 59.0 & 0.0 & 0.0 & 57.2\\
D. table &  13.0 & 71.5 & 18.2 & 0.0 &  7.7 & 21.0\\
toilet &73.2 & 86.9 & 84.2 & 0.0 & 58.3 & 78.5\\
tv & 57.2 & 86.8 & 66.0 & 0.0 & 49.4 & 72.2\\
laptop & 57.2 & 81.7 & 70.0 & 0.0 & 44.0 & 67.9\\
mouse &  68.2 & 86.6 & 78.8 & 44.3 & 81.0 & 62.6\\
remote & 20.7 & 80.1 & 25.8 & 6.8 & 48.8 &  0.0\\
keyboard & 52.4 & 85.2 & 61.5 & 0.0 & 46.8 & 72.2\\
cell phone & 46.1 & 84.9 & 54.3 & 15.0 & 66.2 & 58.1\\
microwave & 61.3 & 91.9 & 66.7 & 0.0 & 60.7 & 94.8\\
oven & 33.3 & 79.1 & 42.1 & 0.0 & 19.5 & 42.4\\
toaster & 0.0 &  0.0 &  0.0 & 0.0 &  0.0 &  0.0\\
sink & 49.5 & 81.8 & 60.5 & 30.8 & 56.8 & 45.7\\
refrigerator & 30.6 & 87.2 & 35.1 & 0.0 & 12.0 & 41.9\\
book & 8.1 & 70.6 & 11.5 & 6.3 & 11.6 & 13.1\\
clock & 59.3 & 86.4 & 68.7 & 40.9 & 68.1 & 92.5\\
vase & 31.8 & 80.5 & 39.4 & 22.4 & 35.3 & 42.5\\
scissors & 0.0 &  0.0 &  0.0 & 0.0 &  0.0 &  0.0\\
teddy bear & 49.0 & 82.4 & 59.4 & 0.0 & 39.8 & 72.8\\
hair drier & 0.0 &  0.0 &  0.0 & 0.0 &  0.0 &  0.0\\
toothbrush & 0.0 &  0.0 &  0.0 & 0.0 &  0.0 &  0.0\\
banner & 5.5 & 79.9 &  6.9 & 0.0 &  0.0 &  0.0\\
blanket & 0.0 &  0.0 &  0.0 & 0.0 &  0.0 &  0.0\\
bridge &  22.0 & 71.3 & 30.8 & 0.0 &  0.0 &  0.0\\
cardboard & 16.6 & 75.7 & 21.9 & 0.0 &  0.0 &  0.0\\
counter & 19.7 & 67.8 & 29.0 & 0.0 &  0.0 &  0.\\
curtain & 45.6 & 83.0 & 54.9 & 0.0 &  0.0 &  0.0\\
door-stuff & 24.4 & 72.8 & 33.6 & 0.0 &  0.0 &  0.0\\
floor-wood & 35.5 & 82.7 & 43.0 & 0.0 &  0.0 &  0.0\\
flower & 12.5 & 65.8 & 19.0 & 0.0 &  0.0 &  0.0\\
fruit & 5.4 & 65.0 &  8.3 & 0.0 &  0.0 &  0.0\\
gravel & 11.6 & 63.5 & 18.2 & 0.0 &  0.0 &  0.0\\
house & 13.5 & 72.5 & 18.6 & 0.0 &  0.0 &  0.0\\
light & 16.1 & 67.5 & 23.8 & 0.0 &  0.0 &  0.0\\
mirror-stuff & 28.2 & 80.4 & 35.1 & 0.0 &  0.0 &  0.0\\
net & 33.7 & 84.3 & 40.0 & 0.0 &  0.0 &  0.0\\
pillow & 0.0 &  0.0 &  0.0 & 0.0 &  0.0 &  0.0\\
platform & 10.3 & 92.5 & 11.1 & 0.0 &  0.0 &  0.0\\
playingfield & 69.4 & 87.6 & 79.2 & 0.0 &  0.0 &  0.0\\
railroad & 25.5 & 72.9 & 35.0 & 0.0 &  0.0 &  0.0\\
river & 22.2 & 82.1 & 27.0 & 0.0 &  0.0 &  0.0\\
road & 45.6 & 83.1 & 54.9 & 0.0 &  0.0 &  0.0\\
roof &  5.2 & 80.8 &  6.5 & 0.0 &  0.0 &  0.0\\
sand & 40.6 & 91.4 & 44.4 & 0.0 &  0.0 &  0.0\\
sea &  71.0 & 91.6 & 77.5 & 0.0 &  0.0 &  0.0\\
shelf & 8.8 & 76.3 & 11.5 & 0.0 &  0.0 &  0.0\\
snow & 81.0 & 91.8 & 88.2 & 0.0 &  0.0 &  0.0\\
stairs &  10.9 & 65.4 & 16.7 & 0.0 &  0.0 &  0.0\\
tent &  5.3 & 53.3 & 10.0 & 0.0 &  0.0 &  0.0\\
towel & 16.8 & 77.7 & 21.6 & 0.0 &  0.0 &  0.0\\
    \hline
    \end{tabular}
  \end{center}
  \caption{Per-class results for COCO dataset. Continued in table~\ref{tab:coco3}}\label{tab:coco2}
\end{table}

\begin{table}[tpb]
  \begin{center}
    \begin{tabular}{|c|c|c|c|c|c|c|}
    \hline
    \textbf{class} & \textbf{PQ} & \textbf{SQ} & \textbf{RQ} & 
    \textbf{PQ$_s$} & \textbf{PQ$_m$} & \textbf{PQ$_l$}\\
    \hline
    \hline
wall-brick & 24.7 & 77.6 & 31.8 & 0.0 &  0.0 &  0.0\\
wall-stone & 10.0 & 92.1 & 10.8 & 0.0 &  0.0 &  0.0\\
wall-tile & 35.2 & 75.7 & 46.5 & 0.0 &  0.0 &  0.0\\
wall-wood & 14.3 & 76.2 & 18.8 & 0.0 &  0.0 &  0.0\\
water-other & 20.9 & 80.3 & 26.1 & 0.0 &  0.0 &  0.0\\
window-blind & 44.6 & 84.7 & 52.6 & 0.0 &  0.0 &  0.0\\
window-other & 22.2 & 73.7 & 30.0 & 0.0 &  0.0 &  0.0\\
tree-merged & 64.6 & 80.7 & 80.0 & 0.0 &  0.0 &  0.0\\
fence-merged & 19.7 & 74.9 & 26.3 & 0.0 &  0.0 &  0.0\\
ceiling-merged & 57.3 & 81.8 & 70.1 & 0.0 &  0.0 &  0.0\\
sky-other-merged & 76.9 & 90.4 & 85.1 & 0.0 &  0.0 &  0.0\\
cabinet-merged & 33.1 & 79.7 & 41.5 & 0.0 &  0.0 &  0.0\\
table-merged & 15.9 & 72.1 & 22.0 & 0.0 &  0.0 &  0.0\\
floor-other-merged & 29.5 & 80.3 & 36.7 & 0.0 &  0.0 &  0.0\\
pavement-merged & 36.4 & 78.9 & 46.2 & 0.0 &  0.0 &  0.0\\
mountain-merged & 39.7 & 76.9 & 51.6 & 0.0 &  0.0 &  0.0\\
grass-merged & 50.3 & 81.2 & 61.9 & 0.0 &  0.0 &  0.0\\
dirt-merged & 27.4 & 77.0 & 35.6 & 0.0 &  0.0 &  0.0\\
paper-merged & 4.7 & 74.6 &  6.3 & 0.0 &  0.0 &  0.0\\
food-other-merged & 14.0 & 78.7 & 17.8 & 0.0 &  0.0 &  0.0\\
building-other-merged & 29.3 & 76.4 & 38.4 & 0.0 &  0.0 &  0.0\\
rock-merged & 31.0 & 78.4 & 39.6 & 0.0 &  0.0 &  0.0\\
wall-other-merged &  45.6 & 79.2 & 57.6 & 0.0 &  0.0 &  0.0\\
rug-merged & 38.3 & 82.7 & 46.4 & 0.0 &  0.0 &  0.0\\

    \hline
    \end{tabular}
  \end{center}
  \caption{Per-class results for COCO dataset. The first 80 classes are from 
the \textit{thing} while the rest 53 are from \textit{stuff} label.}\label{tab:coco3}
\end{table}

\end{document}